\documentclass{article}

\PassOptionsToPackage{numbers,sort&compress}{natbib}
\usepackage[preprint]{neurips_2023}

\usepackage[utf8]{inputenc}
\usepackage[T1]{fontenc}
\usepackage{amsmath,amssymb,amsthm}
\usepackage{graphicx}
\usepackage{float}
\usepackage{hyperref}
\usepackage{url}
\usepackage{algorithm}
\usepackage{algorithmic}
\usepackage{booktabs}
\usepackage{multirow}
\usepackage[normalem]{ulem}

\title{Speedrunning ImageNet Diffusion}
\author{Swayam Bhanded}
\date{}

\begin{document}

\maketitle

\begingroup
\renewcommand{\thefootnote}{}
\footnotetext{Relevant links (code, checkpoints, and experiment logs) are provided in the Resources section.}
\addtocounter{footnote}{-1}
\endgroup

\begin{figure}[H]
\centering
\setlength{\tabcolsep}{0pt}
\renewcommand{\arraystretch}{0}
\begin{tabular}{cccccc}
  \includegraphics[width=0.165\textwidth]{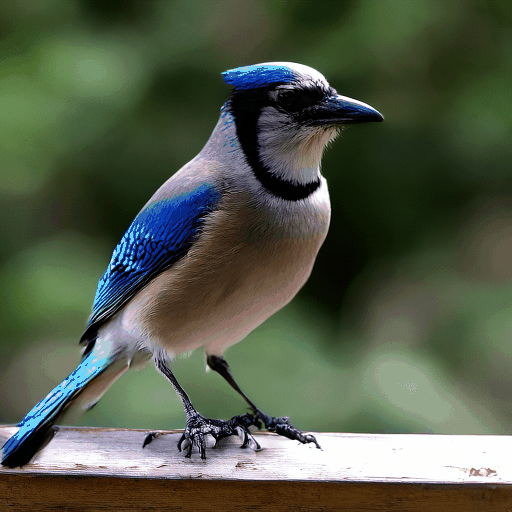} &
  \includegraphics[width=0.165\textwidth]{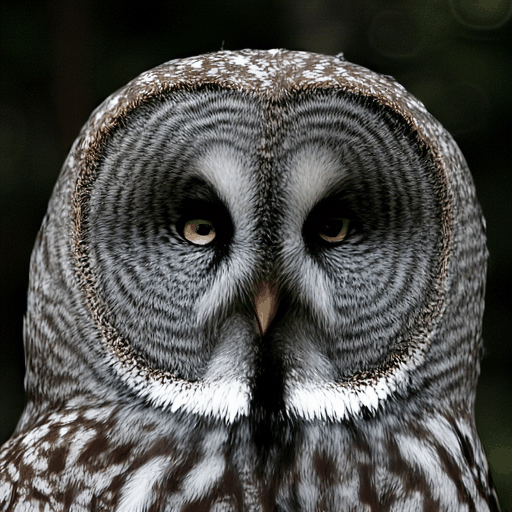} &
  \includegraphics[width=0.165\textwidth]{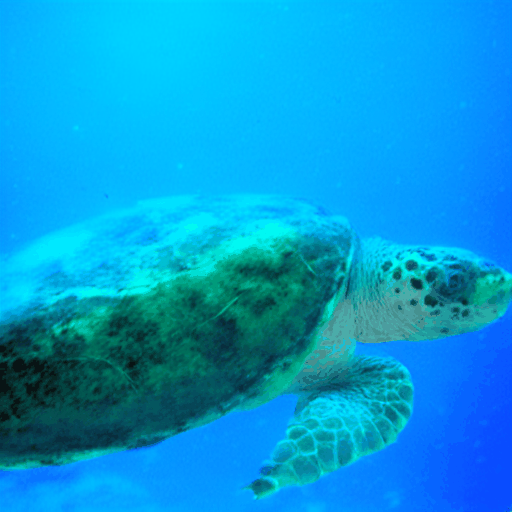} &
  \includegraphics[width=0.165\textwidth]{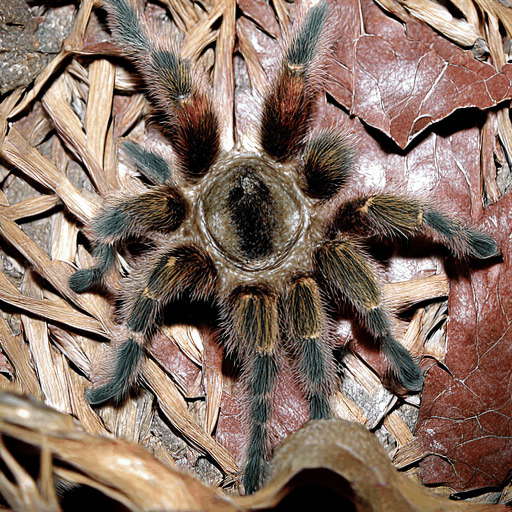} &
  \includegraphics[width=0.165\textwidth]{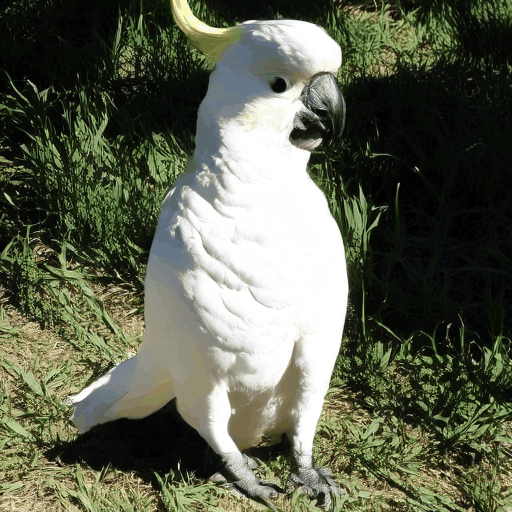} &
  \includegraphics[width=0.165\textwidth]{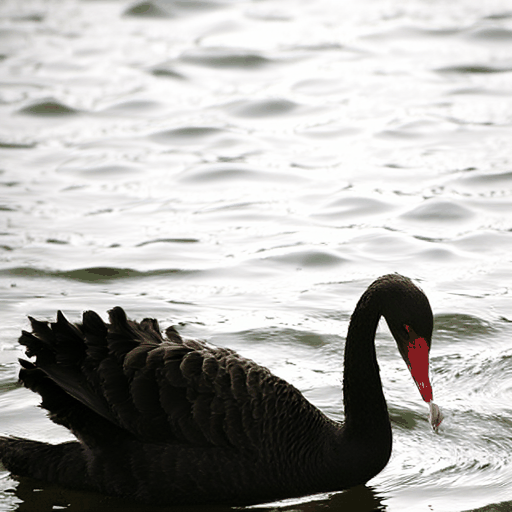} \\
  \includegraphics[width=0.165\textwidth]{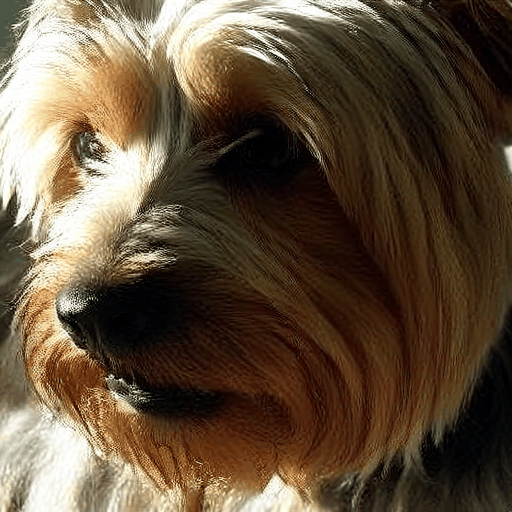} &
  \includegraphics[width=0.165\textwidth]{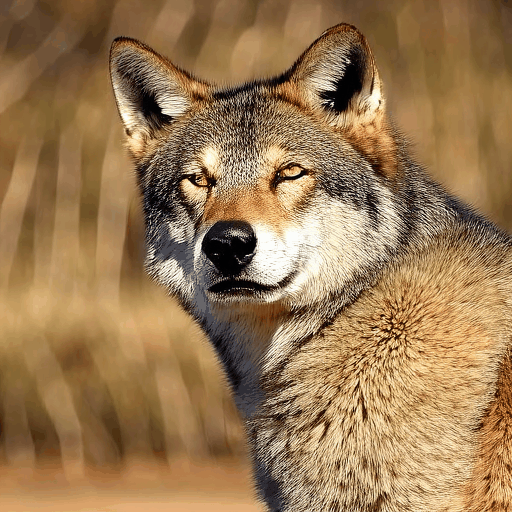} &
  \includegraphics[width=0.165\textwidth]{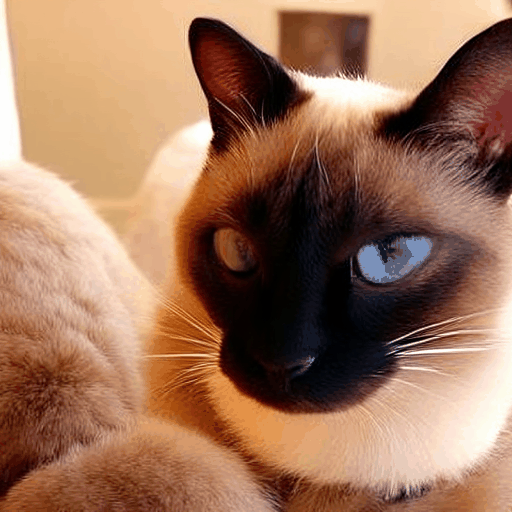} &
  \includegraphics[width=0.165\textwidth]{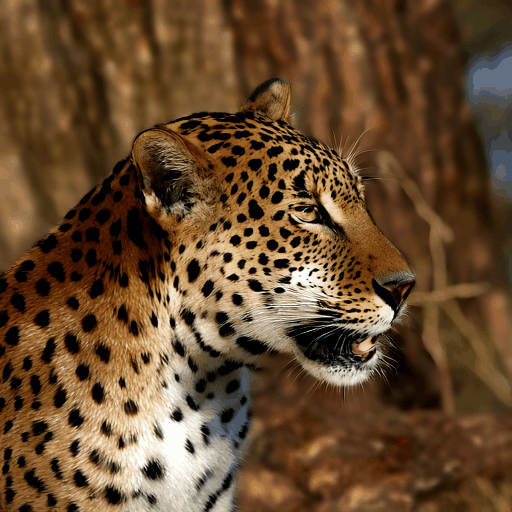} &
  \includegraphics[width=0.165\textwidth]{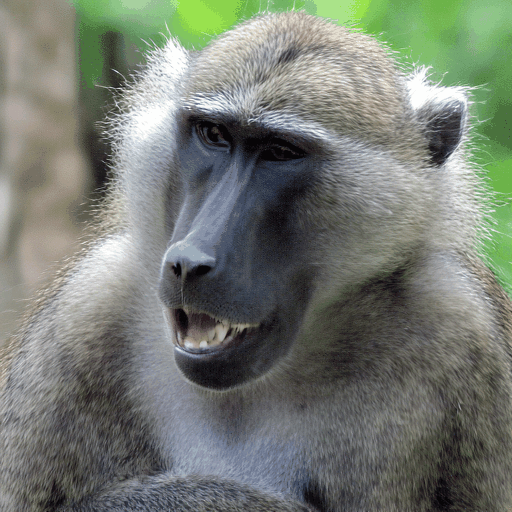} &
  \includegraphics[width=0.165\textwidth]{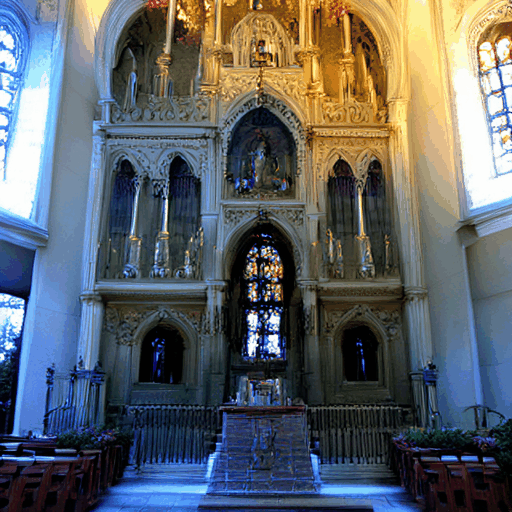} \\
  \includegraphics[width=0.165\textwidth]{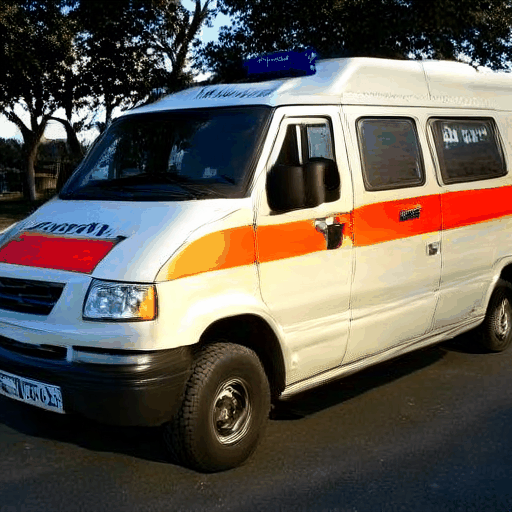} &
  \includegraphics[width=0.165\textwidth]{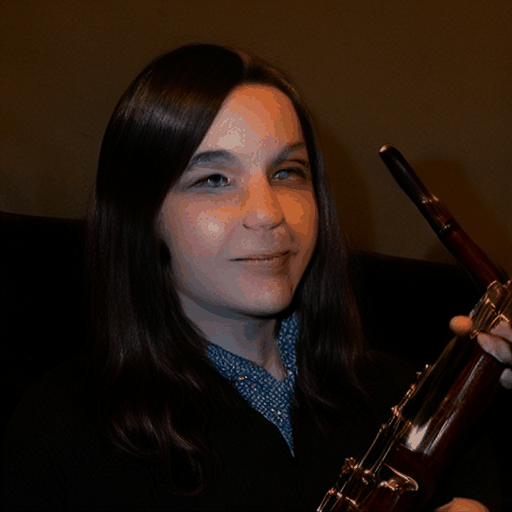} &
  \includegraphics[width=0.165\textwidth]{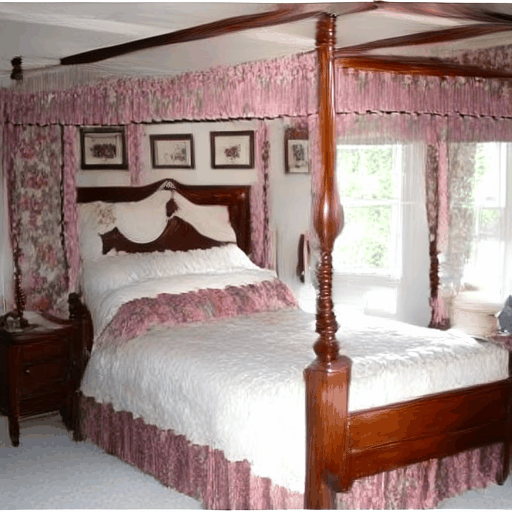} &
  \includegraphics[width=0.165\textwidth]{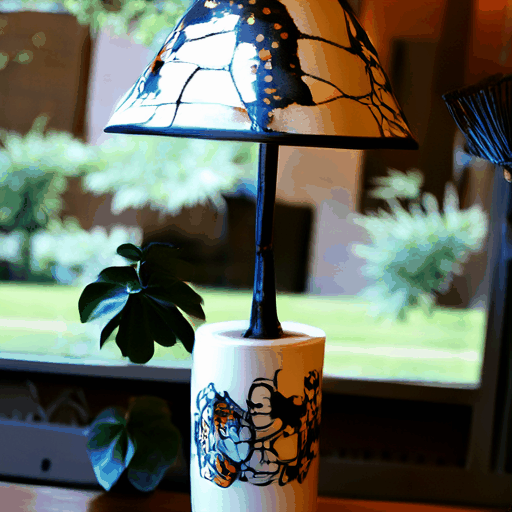} &
  \includegraphics[width=0.165\textwidth]{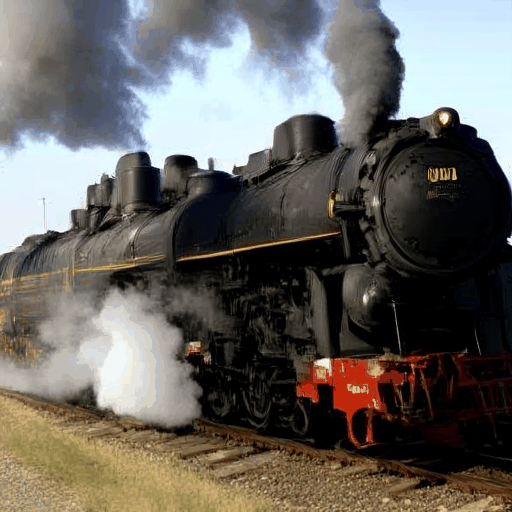} &
  \includegraphics[width=0.165\textwidth]{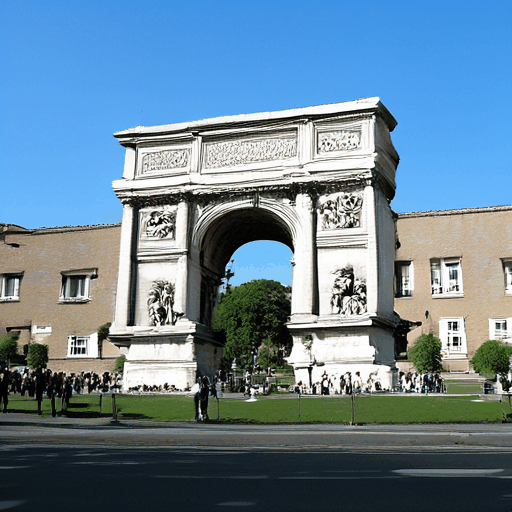} \\
  \includegraphics[width=0.165\textwidth]{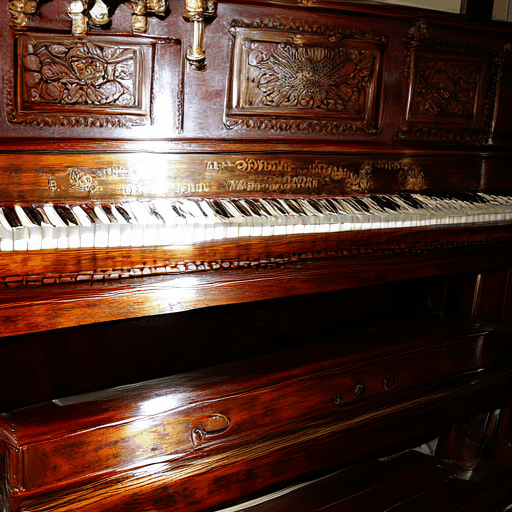} &
  \includegraphics[width=0.165\textwidth]{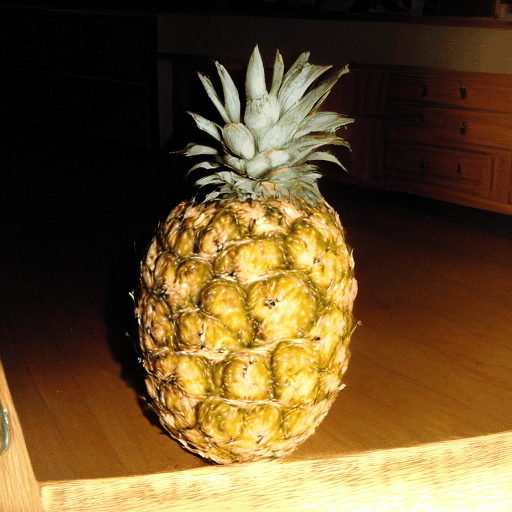} &
  \includegraphics[width=0.165\textwidth]{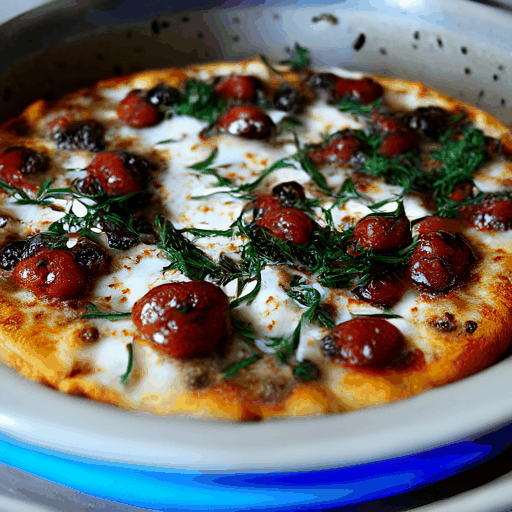} &
  \includegraphics[width=0.165\textwidth]{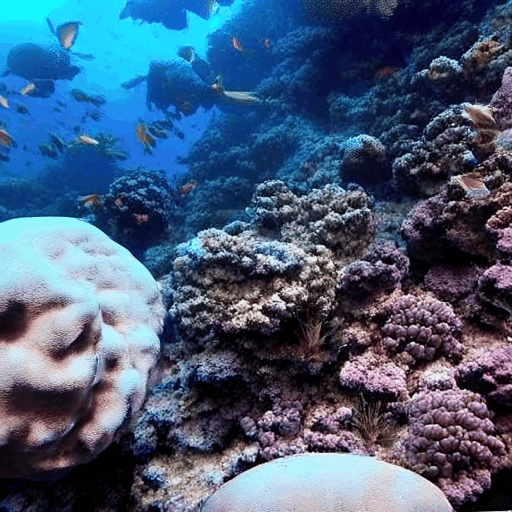} &
  \includegraphics[width=0.165\textwidth]{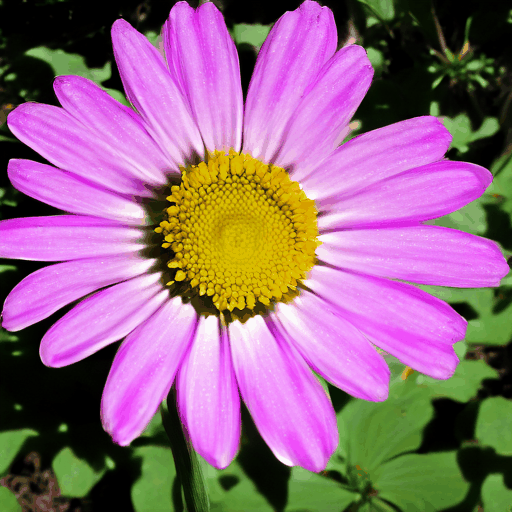} &
  \includegraphics[width=0.165\textwidth]{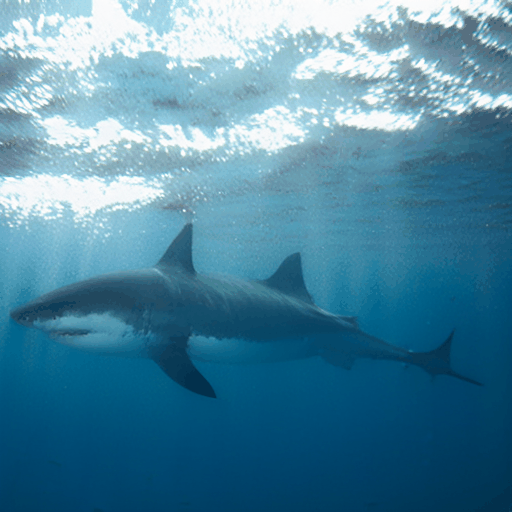}
\end{tabular}
\caption{SR-DiT-B/1 samples on ImageNet-512.}
\label{fig:example_images_grid}
\end{figure}

\begin{abstract}
Recent advances have significantly improved the training efficiency of diffusion transformers. However, these techniques have largely been studied in isolation, leaving unexplored the potential synergies from combining multiple approaches. We present \textbf{SR-DiT} (Speedrun Diffusion Transformer), a framework that systematically integrates token routing, architectural improvements, and training modifications on top of representation alignment. Our approach achieves FID 3.49 and KDD 0.319 on ImageNet-256 using only a 140M parameter model at 400K iterations without classifier-free guidance---comparable to results from 685M parameter models trained significantly longer. To our knowledge, this is a state-of-the-art result at this model size. Through extensive ablation studies, we identify which technique combinations are most effective and document both synergies and incompatibilities. We release our framework as a computationally accessible baseline for future research.
\end{abstract}

\section{Introduction}
\label{sec:introduction}

Diffusion models have emerged as the dominant paradigm for high-quality image generation, yet their training remains computationally expensive. Recent years have witnessed a proliferation of techniques aimed at accelerating diffusion model training: improved architectures~\cite{sprint}, representation alignment~\cite{repa,reg}, better tokenizers~\cite{repae}, and various training modifications. However, these advances have largely been developed and evaluated in isolation, each claiming improvements over increasingly outdated baselines. This fragmented landscape leaves a critical question unanswered: \textit{how do these techniques interact when combined, and what performance is achievable by systematically integrating them?}

The current state of research presents several challenges. First, most techniques are evaluated against vanilla diffusion transformers, ignoring the substantial gains from other concurrent work. Second, flagship results typically require large models (e.g., SiT-XL with 685M parameters) trained for millions of iterations, creating high barriers for academic researchers with limited compute. Third, the interactions between techniques---whether synergistic or redundant---remain poorly understood.

We address these challenges with \textbf{SR-DiT} (Speedrun Diffusion Transformer), a framework that systematically combines recent advances to maximize training efficiency. Our key insight is that many techniques target orthogonal aspects of the training process: representation alignment provides strong learning signals, token routing reduces computational redundancy and improves information flow, modern architectures improve optimization dynamics, and semantic tokenizers provide more learnable latent spaces. By carefully integrating these components, we achieve results that rival or exceed large-scale models while using only a fraction of the compute.

Concretely, starting from the SiT-B/1 architecture (130M parameters), we add improvements to finally achieve FID 3.49 on ImageNet-256 at 400K iterations without classifier-free guidance. For comparison, REG~\cite{reg}---which itself claims 63$\times$ convergence speedup over vanilla SiT---requires SiT-XL (685M parameters) to reach FID 3.4, while REPA~\cite{repa} needs 4 million training steps with SiT-XL to achieve FID 5.9. Our approach thus demonstrates that combining existing techniques intelligently can yield outsized gains, providing a strong and efficient baseline for future research.

Our contributions are:
\begin{itemize}
    \item A systematic study of how recent diffusion training techniques interact when combined, identifying synergies and incompatibilities.
    \item An efficient framework achieving FID 3.49 and KDD 0.319 on ImageNet-256 with only 140M parameters at 400K iterations, comparable to much larger models trained for longer.
    \item Extensive ablations documenting both successful combinations and negative results, providing practical guidance for researchers.
    \item A computationally accessible baseline enabling faster iteration for academic research.
\end{itemize}

\section{Related Work}
\label{sec:related}

\textbf{Diffusion models.} Denoising diffusion probabilistic models~\cite{ddpm,sohl2015deep} and score-based generative models~\cite{song2019generative,song2020score} have become the foundation for state-of-the-art image generation. Flow matching~\cite{lipman2022flow,liu2022flow} provides an alternative formulation with simpler training objectives. The Diffusion Transformer (DiT)~\cite{dit} demonstrated that transformer architectures can match or exceed U-Net performance, while SiT~\cite{sit} extended this to flow matching. We build upon SiT as our base architecture.

\textbf{Representation alignment.} REPA~\cite{repa} introduced the idea of aligning diffusion model hidden states with pretrained vision encoder features, achieving significant training speedups. REG~\cite{reg} extended this with generative objectives, claiming 63$\times$ convergence speedup over vanilla SiT. These methods provide strong learning signals that guide the model toward semantically meaningful representations. We use REG as our starting point and evaluate additional techniques on top.

\textbf{Semantic latent spaces.} The choice of image tokenizer significantly impacts diffusion training dynamics. Standard SD-VAE~\cite{ldm} encodes images into latent spaces optimized for reconstruction, but not necessarily for generation. LightningDiT~\cite{lightningdit} introduced a VAE trained with semantic objectives, producing latent spaces that are more ``diffusable''---easier for diffusion models to learn. INVAE~\cite{repae} (from REPA-E) follows this direction with improved semantic properties. These tokenizers accelerate learning by providing latent representations with stronger semantic structure.

\textbf{Token routing.} TREAD~\cite{tread} demonstrated that routing 50\% of tokens to skip intermediate transformer layers both reduces computation and improves convergence---a counterintuitive finding suggesting that full token processing may be redundant. SPRINT~\cite{sprint} introduces architectural modifications that allow increasing the token drop rate to 75\%, achieving greater efficiency gains. These methods reveal that diffusion transformers have significant computational slack that can be exploited for efficiency.

\textbf{Architectural improvements.} Modern transformer components from language modeling have shown benefits in vision. LightningDiT incorporates SwiGLU activations~\cite{swiglu}, RMSNorm~\cite{rmsnorm}, and RoPE~\cite{rope}. QK normalization~\cite{qknorm} stabilizes attention, and Value Residual Learning~\cite{valueresidual} improves information flow. We evaluate how these architectural choices interact with representation alignment and token routing.

\section{Background}
\label{sec:background}

\subsection{Evaluation Metrics}

While Fr\'echet Inception Distance (FID)~\cite{fid} has been the standard metric for evaluating generative models, recent work has exposed significant limitations in commonly used metrics~\cite{stein2023exposing}. Stein et al. demonstrate that among existing metrics, \textbf{Kernel DINO Distance (KDD)} correlates most strongly with human perceptual judgments. KDD computes distances between generated and real image distributions using DINOv2~\cite{dinov2} features in a kernel-based framework. Lower KDD values indicate better generation quality. We report both KDD and traditional metrics (FID, sFID, IS, Precision, Recall) for comprehensive evaluation.

\subsection{Diffusion / Flow Matching}

We follow SiT~\cite{sit} and train the model with a flow-matching objective using velocity prediction. Given a clean input $x$ (VAE latents) and a timestep $t \in [0,1]$, we construct a noisy sample using an interpolant
\begin{equation}
x_t = \alpha(t)\,x + \sigma(t)\,\epsilon, \qquad \epsilon \sim \mathcal{N}(0, I),
\end{equation}
where $\alpha(t),\sigma(t)$ define the path (e.g., linear or cosine). The corresponding velocity target is
\begin{equation}
v_t^\star = \dot{\alpha}(t)\,x + \dot{\sigma}(t)\,\epsilon.
\end{equation}
The base training loss is mean-squared error on velocity prediction, $\mathbb{E}[\lVert v_\theta(x_t,t) - v_t^\star \rVert_2^2]$.

\subsection{Time Shifting}
\label{sec:time_shifting}

We use time shifting~\cite{logitnormal} to reweight which timesteps the model sees, reducing over-emphasis on high-SNR (easy) denoising steps. We first sample $t \sim \mathcal{U}(0,1)$, then apply the monotone transform
\begin{equation}
t' = \frac{s\,t}{1 + (s-1)t}, \qquad s = \sqrt{\frac{D}{4096}}, \qquad D = C\cdot H\cdot W,
\end{equation}
which (for $s>1$) shifts mass toward larger $t$ (noisier / lower-SNR inputs). We apply the same transformation during both training and sampling. Following Zheng et al.~\cite{rae}, we compute the shift factor from the full latent dimensionality (including channels) and use 4096 as the reference dimension.

\subsection{Rotary Position Embeddings (RoPE)}
\label{sec:rope}

Rotary position embeddings (RoPE)~\cite{rope} encode positional information by rotating query and key vectors in multi-head attention. For a token at position $i$ and head vector $q_i$ (and similarly $k_i$), RoPE rotates each consecutive 2D pair via
\begin{equation}
\mathrm{RoPE}(q_i) = q_i \odot \cos\omega_i + \mathrm{rot}(q_i) \odot \sin\omega_i, \quad \mathrm{rot}([x_{2j},x_{2j+1}]) = [-x_{2j+1},\,x_{2j}],
\end{equation}
where $\omega_i$ are position-dependent frequencies and $\odot$ is elementwise multiplication. For images, we use a 2D extension where positions $i$ correspond to indices on the flattened $H\times W$ patch grid.

\subsection{RMSNorm}

RMSNorm~\cite{rmsnorm} is a normalization layer that scales activations by their root-mean-square (without mean-centering), which reduces computation and can improve stability. For an input vector $x \in \mathbb{R}^d$, RMSNorm computes
\begin{equation}
\mathrm{RMSNorm}(x) = g \odot \frac{x}{\sqrt{\frac{1}{d}\sum_{j=1}^{d} x_j^2 + \epsilon}},
\end{equation}
where $g$ is a learned gain parameter and $\epsilon$ is a small constant.

\subsection{Value Residual Learning}

Value Residual Learning~\cite{valueresidual} modifies attention by injecting a residual connection across the \emph{value} stream. The method caches the value vectors from an early attention block and reuses them as a reference value for subsequent blocks. Let $v^{(\ell)}$ denote the value vectors produced by block $\ell$ (after the value projection) and let $v^{(0)}$ denote the cached reference values (from the first block). We form a mixed value
\begin{equation}
\tilde{v}^{(\ell)} = \lambda\, v^{(0)} + (1-\lambda)\, v^{(\ell)},
\end{equation}
with a learned scalar $\lambda \in [0,1]$ (implemented as a learnable parameter). Attention then uses $\tilde{v}^{(\ell)}$ in place of $v^{(\ell)}$.

\subsection{Token Routing and Path-Drop Guidance}

TREAD~\cite{tread} and SPRINT~\cite{sprint} implement token routing by temporarily dropping a large fraction of tokens in the middle transformer blocks: a dense prefix processes all tokens, then only the retained (sparse) tokens are propagated through a sequence of mid blocks, and the dropped tokens are reintroduced for the final blocks. SPRINT improves this reintroduction step by padding the sparse sequence back to the full length with a learned \texttt{[MASK]} token and explicitly fusing the padded sparse stream with the dense stream (e.g., concatenation followed by a projection).

We also use \emph{path-drop guidance} (PDG), a CFG-style heuristic in which the unconditional prediction is intentionally weakened by skipping the routed mid blocks entirely (i.e., dropping the sparse path). Let $v_\theta(x_t,t,c)$ denote the conditional model prediction under class conditioning $c$ (sparse path enabled) and $v_\theta^{\text{weak}}(x_t,t)$ the unconditional prediction computed with the routed mid blocks skipped. The guided prediction is
\begin{equation}
v_\theta^{\text{guide}}(x_t,t,c) = v_\theta^{\text{weak}}(x_t,t) + s\left(v_\theta(x_t,t,c) - v_\theta^{\text{weak}}(x_t,t)\right),
\end{equation}
where $s$ is the guidance scale. Importantly, we use PDG \textbf{only for qualitative sampling} (visualizations) and \textbf{do not} use PDG for the quantitative metrics reported in the paper.

\subsection{Contrastive Flow Matching}

Contrastive Flow Matching (CFM)~\cite{cfm} introduces an additional training objective that improves convergence speed. The CFM loss is computed by contrasting model outputs with random targets:
\begin{equation}
\mathcal{L}_{\text{CFM}} = -\lambda \mathbb{E}\left[\left\|v_\theta(x_t, t) - \hat{v}_{\text{target}}\right\|^2\right]
\end{equation}
where $\hat{v}_{\text{target}}$ is a random training target unrelated to $x_t$ (obtained by shuffling the minibatch of the velocity target elementwise), and $\lambda$ is a weighting coefficient (default 0.05). The negative sign encourages the model to maximize distance from other samples' predictions.

\subsection{Representation Alignment (REPA and REG)}

We build on Representation Alignment for Generation (REPA)~\cite{repa} and its extension REG~\cite{reg}, which add an auxiliary representation objective to the standard denoising / flow-matching loss.
We use the same noising process and velocity-prediction objective described above.

For the representation targets, we pass the \emph{clean} image through a frozen vision encoder (DINOv2~\cite{dinov2}) to obtain per-token targets $z$ (and a global CLS embedding $c_{\text{cls}}$). The diffusion transformer produces intermediate hidden states that are mapped through small MLP projectors to predicted representations $\tilde{z}$. REPA uses a projection loss based on cosine similarity:
\begin{equation}
\mathcal{L}_{\text{REPA}} = -\lambda_{\text{REPA}}\,\frac{1}{M}\sum_{m=1}^{M} \left\langle \frac{z_m}{\lVert z_m \rVert_2},\, \frac{\tilde{z}_m}{\lVert \tilde{z}_m \rVert_2} \right\rangle,
\end{equation}
where $m$ indexes tokens and we use $\lambda_{\text{REPA}}=0.5$ by default.

REG extends this setup by also diffusing the DINO CLS embedding alongside the latents.

\section{Method}
\label{sec:method}

Our approach systematically combines recent advances to maximize training efficiency. We start from REG~\cite{reg} with INVAE~\cite{repae} as our baseline, then progressively add architectural improvements and training objective modifications.

\subsection{Architectural Improvements}

We evaluate several modern transformer components that have shown benefits in language and vision models:

\textbf{SPRINT}~\cite{sprint}: We use SPRINT token routing with a drop ratio of 0.75 (keeping 25\% tokens) in the sparse path. Following the standard SPRINT split, we run 2 dense ``encoder'' blocks on all tokens, route over the middle blocks (operating on the sparse subset, then padding back and fusing), and run 2 dense ``decoder'' blocks on the fused full sequence.

\textbf{RMSNorm}~\cite{rmsnorm}: Starting from our fork of the REG baseline, we replace all LayerNorm instances in the backbone with RMSNorm. Concretely, this includes: (i) the two per-block normalizations before attention and before the MLP in every transformer block (\texttt{norm1}, \texttt{norm2}); (ii) the per-head query/key normalizers used by QK Norm (\texttt{q\_norm}, \texttt{k\_norm}); (iii) the final normalization before the output projection (\texttt{norm\_final}); and (iv) the RMSNorm applied to the REG-diffused CLS embedding before concatenation (\texttt{wg\_norm}).

\textbf{Rotary Position Embeddings (RoPE)}~\cite{rope}: We use 2D RoPE (see Section~\ref{sec:rope}) via the EVA-02-style implementation~\cite{eva02}, later adapted by the TREAD authors to support routed / sparse token subsets for their LightningDiT+TREAD experiments~\cite{lightningdit,tread}.\footnote{\url{https://github.com/flixmk/LightningDiT_TREAD/blob/main/models/pos_embed_tread.py}}
In routed blocks, we pass per-token indices \texttt{rope\_ids} so each retained token is rotated using its original spatial position. We further modify the implementation to exclude the leading class token (the REG-diffused CLS embedding) from rotation.

\textbf{QK Normalization}~\cite{qknorm}: Normalizing query and key vectors before computing attention scores stabilizes training dynamics.

\textbf{Value Residual Learning}~\cite{valueresidual}: Adding a residual connection around the value projection improves gradient flow and model expressiveness.

These components were proposed independently in various contexts; our contribution is evaluating their interactions when combined with representation alignment and token routing.

\subsection{Training Objective}

We incorporate Contrastive Flow Matching (CFM)~\cite{cfm}, which adds an auxiliary loss that improves convergence speed (see Section~\ref{sec:background}). Overall, our training loss is
\begin{equation}
\mathcal{L} = \mathcal{L}_{\text{vel}} + \lambda_{\text{REPA}}\,\mathcal{L}_{\text{REPA}} + \lambda_{\text{cls}}\,\mathcal{L}_{\text{cls}} + \lambda_{\text{CFM}}\,\mathcal{L}_{\text{CFM}},
\end{equation}
where $\mathcal{L}_{\text{vel}}$ is the standard velocity-prediction MSE on latents (Section~\ref{sec:background}), $\mathcal{L}_{\text{REPA}}$ is the projection loss (Section~\ref{sec:background}), $\mathcal{L}_{\text{cls}}$ is the velocity-prediction MSE for the REG-diffused CLS embedding, and $\mathcal{L}_{\text{CFM}}$ is the contrastive term. We use $\lambda_{\text{REPA}}=0.5$, $\lambda_{\text{cls}}=0.03$, and $\lambda_{\text{CFM}}=0.05$.

\subsection{Time Shifting}

We apply time shifting (Section~\ref{sec:time_shifting}) during both training and sampling.

\section{Experiments}
\label{sec:experiments}

\subsection{Experimental Setup}

We conduct experiments on ImageNet-256~\cite{imagenet}. We build upon the SiT-B~\cite{sit} and REG~\cite{reg} architecture and compare our ablations against multiple baselines. Our model uses the SiT-B/1 architecture (patch size 1 instead of 2), starting with 132M parameters for the baseline, as INVAE has 16$\times$ spatial compression compared to SD-VAE~\cite{ldm}'s 8$\times$ compression. SPRINT modifications increase the model to 140M parameters. This architecture choice significantly reduces computational costs compared to larger models like SiT-XL (685M parameters) while maintaining strong performance.

 Training the final SR-DiT-B/1 architecture to 400K iterations took approximately 10 hours on a single node with 8$\times$ NVIDIA H200 GPUs for ImageNet-256 (80 GPU-hours), and approximately 15 hours for ImageNet-512 (120 GPU-hours).

We evaluate generation quality using Kernel DINO Distance (KDD) as our primary metric, along with standard metrics: Fr\'echet Inception Distance (FID)~\cite{fid}, spatial FID (sFID)~\cite{sfid}, Inception Score (IS)~\cite{inception_score}, Precision, and Recall~\cite{precision_recall}. All metrics are computed on 50K generated samples. Following RAE~\cite{rae}, we use balanced label sampling during generation for metric calculation, ensuring each class is equally represented in the generated samples. This corrects an implementation detail where random sampling can lead to imbalanced class distributions, improving FID marginally. \textbf{All results are reported without CFG or PDG}.

\subsection{Results}

We use REG with INVAE as our base configuration and systematically evaluate architectural and training improvements. Table~\ref{tab:main_results} presents our main results at 400K iterations, comparing against SiT-B/2 baselines.

\begin{table}[t]
\centering
\caption{Performance comparison on ImageNet-256 at 400K iterations. Lower FID/sFID/KDD and higher IS/Precision/Recall are better. All methods evaluated at NFE=250 without CFG or PDG.}
\label{tab:main_results}
\small
\begin{tabular}{l@{\hspace{0.5em}}c@{\hspace{0.5em}}c@{\hspace{0.5em}}c@{\hspace{0.5em}}c@{\hspace{0.5em}}c@{\hspace{0.5em}}c@{\hspace{0.5em}}c}
\toprule
Method & \#Params & FID$\downarrow$ & sFID$\downarrow$ & IS$\uparrow$ & Prec.$\uparrow$ & Rec.$\uparrow$ & KDD$\downarrow$ \\
\midrule
\multicolumn{8}{l}{\textit{Baselines (SiT-B/2, SD-VAE)}} \\
SiT-B/2 & 130M & 33.0 & 6.46 & 43.7 & 0.53 & 0.63 & -- \\
+ REPA & 130M & 24.4 & 6.40 & 59.9 & 0.59 & 0.65 & -- \\
+ REG & 132M & 15.2 & 6.69 & 94.6 & 0.64 & 0.63 & -- \\
\midrule
\multicolumn{8}{l}{\textit{SR-DiT (SiT-B/1, INVAE)}} \\
REG + INVAE & 132M & 10.56 & \uline{5.49} & 104.5 & 0.691 & \textbf{0.632} & 0.586 \\
+ SPRINT & 140M & 4.58 & 7.04 & 188.7 & 0.760 & 0.562 & 0.385 \\
+ RMSNorm & 140M & 4.56 & 6.58 & 186.8 & 0.772 & 0.562 & 0.381 \\
+ RoPE & 140M & 4.09 & 6.22 & 194.2 & 0.778 & \uline{0.563} & 0.368 \\
+ QK Norm & 140M & 4.02 & 6.18 & 193.5 & 0.784 & \uline{0.563} & 0.364 \\
+ Value Residual & 140M & 3.64 & 5.98 & 202.0 & 0.798 & 0.560 & 0.353 \\
+ CFM & 140M & 3.61 & \textbf{5.26} & \uline{211.9} & \textbf{0.817} & 0.536 & \textbf{0.319} \\
+ Time Shifting & 140M & \uline{3.56} & 5.64 & 209.6 & 0.807 & 0.550 & 0.336 \\
+ Balanced Sampling & 140M & \textbf{3.49} & 5.67 & \textbf{221.2} & \uline{0.808} & 0.546 & \uline{0.332} \\
\bottomrule
\end{tabular}
\end{table}

In addition to ImageNet-256, we evaluate SR-DiT-B/1 on ImageNet-512 using the same training setup scaled to $512\times512$ resolution. We compare against the DiT-XL/2 and U-DiT-B baselines reported in U-DiTs~\cite{tian2024udits} as these are the only reported results we could find for FID-50k on ImageNet-512 at 400k iterations. As shown in Table~\ref{tab:imagenet512}, SR-DiT-B/1 achieves strong ImageNet-512 performance. All results are evaluated without CFG or PDG.

\begin{table}[t]
\centering
\caption{Performance comparison on ImageNet-512 at 400K iterations. Baseline results for DiT-XL/2* and U-DiT-B are taken from U-DiTs~\cite{tian2024udits}. Lower FID/sFID/KDD and higher IS/Precision/Recall are better.}
\label{tab:imagenet512}
\small
\begin{tabular}{l@{\hspace{0.5em}}c@{\hspace{0.5em}}c@{\hspace{0.5em}}c@{\hspace{0.5em}}c@{\hspace{0.5em}}c@{\hspace{0.5em}}c}
\toprule
Method & FID$\downarrow$ & sFID$\downarrow$ & IS$\uparrow$ & Prec.$\uparrow$ & Rec.$\uparrow$ & KDD$\downarrow$ \\
\midrule
DiT-XL/2* & 20.94 & 6.78 & 66.30 & 0.745 & 0.581 & -- \\
U-DiT-B & 15.39 & 6.86 & 92.73 & 0.756 & \textbf{0.605} & -- \\
\midrule
SR-DiT-B/1 (ours) & \textbf{4.23} & \textbf{5.46} & \textbf{192.34} & \textbf{0.839} & 0.531 & \textbf{0.306} \\
\bottomrule
\end{tabular}
\end{table}

\begin{figure}[t]
\centering
\includegraphics[width=0.85\textwidth]{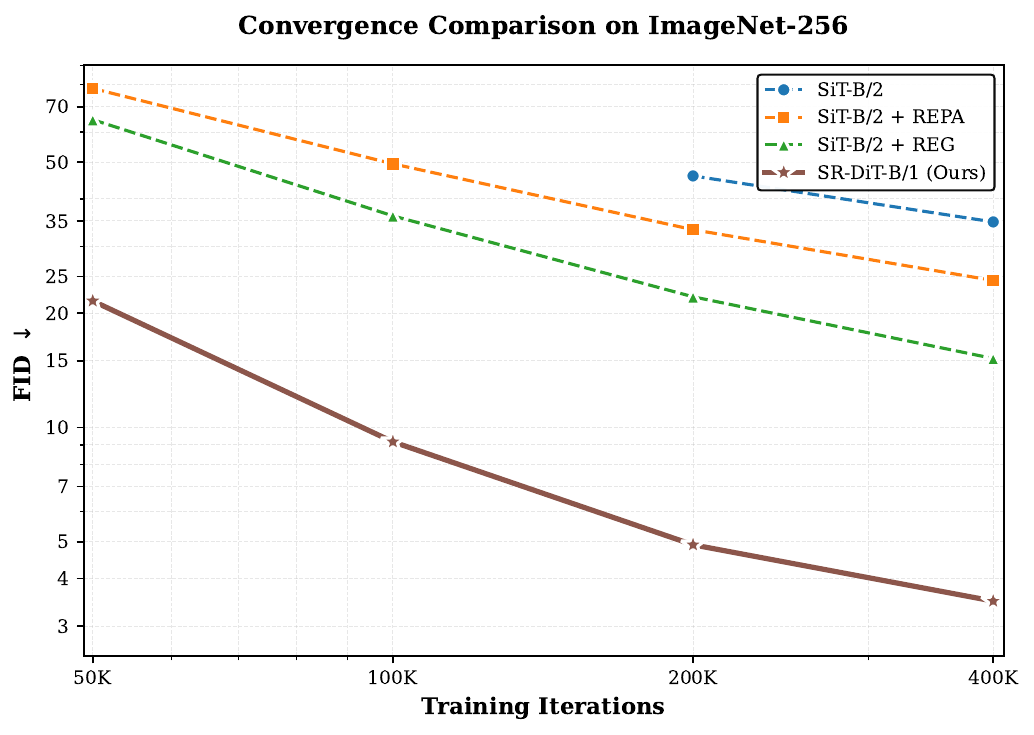}
\caption{Training convergence comparison on ImageNet-256. SR-DiT-B/1 achieves strong performance with substantial convergence speedup.}
\label{fig:convergence}
\end{figure}

\begin{figure}[t]
\centering
\setlength{\tabcolsep}{1pt}
\begin{tabular}{cccccc}
    \includegraphics[width=0.15\textwidth]{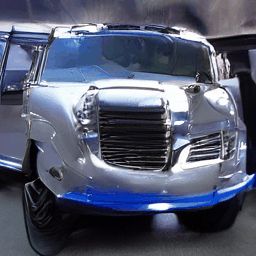} &
    \includegraphics[width=0.15\textwidth]{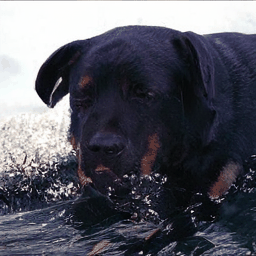} &
    \includegraphics[width=0.15\textwidth]{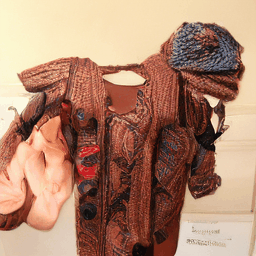} &
    \includegraphics[width=0.15\textwidth]{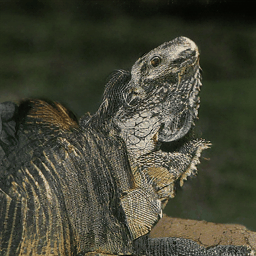} &
    \includegraphics[width=0.15\textwidth]{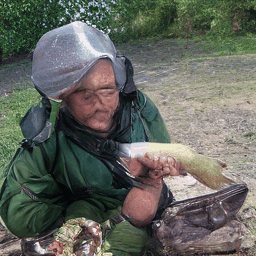} &
    \includegraphics[width=0.15\textwidth]{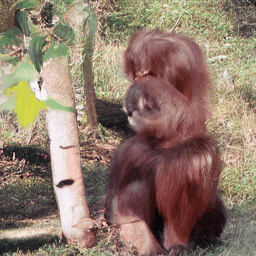} \\
    \includegraphics[width=0.15\textwidth]{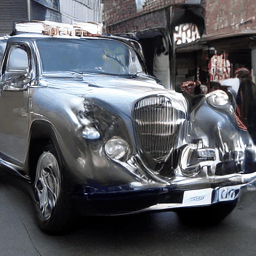} &
    \includegraphics[width=0.15\textwidth]{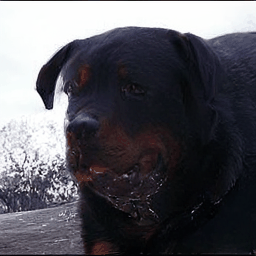} &
    \includegraphics[width=0.15\textwidth]{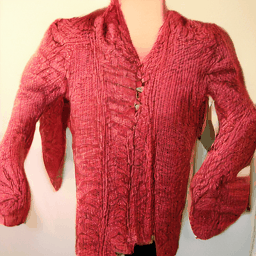} &
    \includegraphics[width=0.15\textwidth]{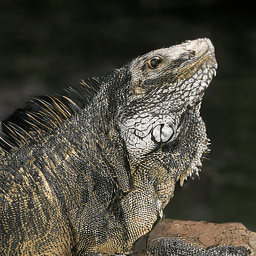} &
    \includegraphics[width=0.15\textwidth]{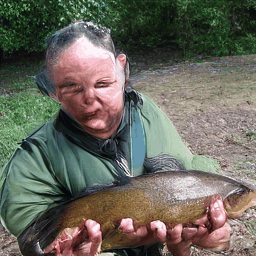} &
    \includegraphics[width=0.15\textwidth]{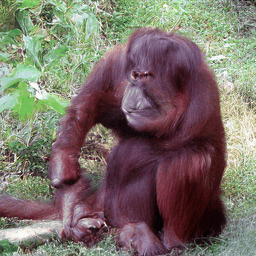}
\end{tabular}
\caption{Qualitative comparison between our REG + INVAE starting point (top row) and the final SR-DiT-B/1 architecture (bottom row) on ImageNet-256 at 400K training iterations. Generated without CFG or PDG, using the same random seed and class label for both models.}
\label{fig:qualitative_reg_vs_srd}
\end{figure}

Figure~\ref{fig:qualitative_reg_vs_srd} qualitatively illustrates how our final architecture sharpens details and improves semantic fidelity over the REG + INVAE baseline under identical sampling conditions.

\textbf{Baseline comparison.} The original SiT-B/2~\cite{sit} with 130M parameters requires 400K iterations to achieve FID 33.0. REPA~\cite{repa} improves this to FID 24.4, while REG~\cite{reg} reaches FID 15.2 at 400K iterations with 132M parameters. Our baseline configuration (REG + INVAE) with SiT-B/1 achieves FID 10.56 using 132M parameters, already surpassing REG with SD-VAE at the same parameter count.

\subsection{Analysis}

\textbf{What matters most.} The dominant improvement comes from representation alignment and entanglement (REPA/REG), token routing (SPRINT), and from using a semantic VAE (INVAE)

\textbf{Other components help cumulatively.} RMSNorm, RoPE, QK Norm, and Value Residual are individually modest but consistently beneficial in our setting, improving optimization stability and information flow when stacked.

\section{Discussion}
\label{sec:discussion}

\subsection{Limitations and Future Work}

While our results demonstrate strong improvements, several avenues remain for future exploration:

\textbf{Scaling to larger models.} While we focus on the efficient SiT-B/1 architecture, investigating whether these improvements transfer to larger models (SiT-L, SiT-XL) would be valuable for understanding scalability.

\textbf{Text to image generation.} Extending these techniques to text-to-image generation would demonstrate generalization and practical utility.

\textbf{Further optimizations.} There are numerous other techniques which we did not explore due to a lack of time and resources. Future work can investigate these techniques to further improve performance.

\section{Conclusion}
\label{sec:conclusion}

We present SR-DiT, a framework that achieves efficient diffusion model training by combining representation alignment with modern architectural improvements and training modifications. Starting from REG with INVAE (132M parameters), we demonstrate that progressive modifications (SPRINT, RMSNorm, RoPE, QK Norm, Value Residual, CFM, Time Shifting, Balanced Sampling) create strong synergies, achieving FID 3.49 and KDD 0.319 on ImageNet-256 with the efficient SiT-B/1 architecture (140M parameters). We also adopt KDD as a more reliable evaluation metric. Our systematic ablations identify which components contribute most to performance gains, providing insights for future work in efficient diffusion model training.

\section*{Resources}

Code, checkpoints, and experiment logs are available at:
\begin{itemize}
    \item \textbf{GitHub (final SR-DiT code)}: \url{https://github.com/SwayStar123/SpeedrunDiT}
    \item \textbf{Checkpoints}: \url{https://huggingface.co/SwayStar123/SpeedrunDiT/tree/main}
    \item \textbf{Weights \& Biases runs}: \url{https://wandb.ai/kagaku-ai/REG/}
\end{itemize}
Original code for the full set of ablations can be found as branches in \url{https://github.com/SwayStar123/REG}; the final architecture was consolidated into the SpeedrunDiT repository.

\begin{ack}
We gratefully acknowledge support from WayfarerLabs (\url{https://wayfarerlabs.ai}), also known as Open World Labs, for sponsoring the compute resources used in this work.
\end{ack}

\bibliographystyle{plainnat}
\bibliography{references}

\appendix

\section{Additional Results}

This appendix documents additional quantitative results that complement the main text.

\subsection{Intermediate SR-DiT-B/1 Training Checkpoints}

Table~\ref{tab:intermediate_srd} reports SR-DiT-B/1 performance at intermediate training checkpoints, along with the baseline diffusion transformers and the final SR-DiT-B/1 model at 400K iterations.

\begin{table}[h]
\centering
\caption{SR-DiT performance at intermediate training checkpoints on ImageNet-256. Lower FID/sFID/KDD and higher IS/Precision/Recall are better. All methods evaluated at NFE=250 without CFG or PDG.}
\label{tab:intermediate_srd}
\small
\begin{tabular}{l@{\hspace{0.75em}}c@{\hspace{0.75em}}c@{\hspace{0.75em}}c@{\hspace{0.75em}}c@{\hspace{0.75em}}c@{\hspace{0.75em}}c@{\hspace{0.75em}}c}
\toprule
Method & Iter. & FID$\downarrow$ & sFID$\downarrow$ & IS$\uparrow$ & Prec.$\uparrow$ & Rec.$\uparrow$ & KDD$\downarrow$ \\
\midrule
SiT-B/2 & 400K & 33.0 & 6.46 & 43.7 & 0.530 & 0.630 & -- \\
+ REPA & 400K & 24.4 & 6.40 & 59.9 & 0.590 & \textbf{0.650} & -- \\
+ REG & 400K & 15.2 & 6.69 & 94.6 & 0.640 & 0.630 & -- \\
\midrule
SR-DiT-B/1 & 50K & 21.55 & 7.75 & 61.37 & 0.677 & 0.475 & 0.926 \\
SR-DiT-B/1 & 100K & 9.17 & 6.52 & 124.69 & 0.767 & 0.484 & 0.584 \\
SR-DiT-B/1 & 200K & 4.91 & 5.95 & 184.05 & 0.801 & 0.523 & 0.408 \\
SR-DiT-B/1 (final) & 400K & \textbf{3.49} & \textbf{5.49} & \textbf{221.2} & \textbf{0.808} & 0.546 & \textbf{0.332} \\
\bottomrule
\end{tabular}
\end{table}

\section{Additional Ablations}
\label{sec:appendix}

This section documents ablation experiments that did not yield improvements, providing insights into which techniques are less effective when combined with representation alignment.

\subsection{Alternative Activations}

We evaluated replacing standard feedforward activations with SwiGLU~\cite{swiglu}, as proposed in LightningDiT. Table~\ref{tab:swiglu_ablation} shows results at 400K iterations.

\begin{table}[h]
\centering
\caption{SwiGLU activation ablation at 400K iterations. Both experiments use REG + INVAE + SPRINT (140M parameters). SwiGLU shows marginal improvement with noticeable training slowdown.}
\label{tab:swiglu_ablation}
\small
\begin{tabular}{l@{\hspace{1em}}c@{\hspace{1em}}c@{\hspace{1em}}c@{\hspace{1em}}c@{\hspace{1em}}c@{\hspace{1em}}c}
\toprule
Method & FID$\downarrow$ & sFID$\downarrow$ & IS$\uparrow$ & Prec.$\uparrow$ & Rec.$\uparrow$ & KDD$\downarrow$ \\
\midrule
SPRINT (baseline) & \textbf{4.58} & 7.04 & \textbf{188.7} & \textbf{0.760} & 0.562 & \textbf{0.385} \\
+ SwiGLU & 4.65 & \textbf{6.84} & 185.1 & 0.759 & \textbf{0.568} & \textbf{0.385} \\
\bottomrule
\end{tabular}
\end{table}

SwiGLU showed marginal differences in performance metrics. While sFID improved slightly (6.84 vs 7.04) and KDD remained identical (0.385), FID slightly degraded (4.65 vs 4.58) and IS decreased (185.1 vs 188.7). Given these mixed results and a noticeable slowdown in training speed, we excluded SwiGLU from our final framework. The minimal performance differences did not justify the computational overhead.

We also tested additional activation variants. Table~\ref{tab:activation_variants} shows results at 400K iterations.

\begin{table}[h]
\centering
\caption{Alternative activation ablations at 400K iterations. All experiments use REG + INVAE + SPRINT + RMSNorm + RoPE + QK Norm. Standard GELU achieves the best compatibility with Value Residual.}
\label{tab:activation_variants}
\small
\begin{tabular}{l@{\hspace{1em}}c@{\hspace{1em}}c@{\hspace{1em}}c@{\hspace{1em}}c@{\hspace{1em}}c@{\hspace{1em}}c}
\toprule
Activation & FID$\downarrow$ & sFID$\downarrow$ & IS$\uparrow$ & Prec.$\uparrow$ & Rec.$\uparrow$ & KDD$\downarrow$ \\
\midrule
GELU (baseline) & 4.02 & \textbf{6.18} & 193.5 & \textbf{0.784} & 0.563 & 0.364 \\
RELU$^2$ & \textbf{3.81} & 6.47 & \textbf{198.4} & 0.778 & \textbf{0.572} & \textbf{0.351} \\
XieLU & 4.10 & 6.35 & 191.4 & 0.783 & 0.562 & 0.357 \\
Lopsided Leaky RELU$^2$ & 4.11 & 6.67 & 189.4 & 0.768 & 0.569 & 0.366 \\
\bottomrule
\end{tabular}
\end{table}

\textbf{RELU$^2$ incompatibility with Value Residual.} We tested RELU$^2$~\cite{primer}, which initially showed strong results (FID 3.81, KDD 0.351). However, when combined with Value Residual Learning, performance degraded significantly (FID 3.81 $\to$ 3.81, IS 198.4 $\to$ 195.1, KDD 0.351 $\to$ 0.355). In contrast, standard GELU with Value Residual achieved superior results (FID 3.64, IS 202.0, KDD 0.353). Since Value Residual provides larger gains than RELU$^2$ alone, we adopted GELU as our activation function. This incompatibility suggests that RELU$^2$ and Value Residual may optimize overlapping aspects of the model, leading to diminishing returns when combined.

\textbf{XieLU.} We tested XieLU~\cite{xielu}, a recently proposed activation function derived through integration principles. While XieLU achieved competitive results (FID 4.10, KDD 0.357), it underperformed RELU$^2$ across most metrics and exhibited noticeably slower training speed. The performance gap and computational overhead made it unsuitable for our framework.

\textbf{Lopsided Leaky RELU$^2$.} We tested a variant of RELU$^2$ with negative slope handling, defined as:
\begin{equation}
f(x) = \begin{cases}
x^2 & \text{if } x > 0 \\
\alpha x & \text{if } x \leq 0
\end{cases}
\end{equation}
where we use $\alpha = 0.01$. While this variant showed marginally better training loss (approximately 0.0005 lower) than standard RELU$^2$, evaluation metrics were noticeably worse across all measures (FID 4.11 vs 3.81, KDD 0.366 vs 0.351). This discrepancy indicates the variant is prone to overfitting. We document this negative result because multiple researchers have independently experimented with this exact variation without success, yet no published work references it. By publishing this result, we hope to prevent others from expending computational resources on this unpromising direction.

\subsection{Dispersive Loss}

The dispersive loss~\cite{wang2024diffuse} for improving representation diversity yielded negligible performance differences compared to our baseline, confirming that representation alignment from REG already provides sufficient diversity in the learned representations.

\subsection{SARA Structural Loss}

We evaluated SARA's autocorrelation-based structural loss~\cite{sara}, which encourages structural coherence in generated images. Table~\ref{tab:sara_ablation} shows results at 400K iterations with different loss weights. The adversarial component of SARA caused training instability and was excluded.

\begin{table}[h]
\centering
\caption{SARA structural loss ablation at 400K iterations. All experiments use REG + INVAE + SPRINT + RMSNorm + RoPE + QK Norm + Value Residual. The structural loss does not improve upon the baseline.}
\label{tab:sara_ablation}
\small
\begin{tabular}{l@{\hspace{1em}}c@{\hspace{1em}}c@{\hspace{1em}}c@{\hspace{1em}}c@{\hspace{1em}}c@{\hspace{1em}}c}
\toprule
Method & FID$\downarrow$ & sFID$\downarrow$ & IS$\uparrow$ & Prec.$\uparrow$ & Rec.$\uparrow$ & KDD$\downarrow$ \\
\midrule
Baseline & \textbf{3.64} & \textbf{5.98} & \textbf{202.0} & \textbf{0.798} & 0.560 & 0.353 \\
+ SARA ($\lambda=0.5$) & 3.71 & 6.07 & 200.1 & 0.793 & 0.553 & \textbf{0.351} \\
+ SARA ($\lambda=0.25$) & 3.80 & 6.00 & 195.6 & 0.786 & \textbf{0.563} & 0.361 \\
\bottomrule
\end{tabular}
\end{table}

At the default weight ($\lambda=0.5$), most metrics degraded slightly despite a marginal KDD improvement. Reducing the weight to $\lambda=0.25$ further degraded performance across most metrics.

\subsection{Alternative Training Objectives}

We evaluated alternative training objectives beyond standard flow matching to determine if they could improve generation quality or training efficiency.

\textbf{Time-Weighted Contrastive Flow Matching (TCFM).} We hypothesized that CFM's contrastive loss might be detrimental at low noise levels, where it could perturb the learned flow even when the clean image structure is already well-defined. We proposed Time-Weighted CFM (TCFM), which reduces the CFM influence as samples approach the clean image:
\begin{equation}
\mathcal{L}_{\text{TCFM}} = -t \cdot \lambda \mathbb{E}\left[\left\|v_\theta(x_t, t) - \hat{v}_{\text{target}}\right\|^2\right]
\end{equation}
where $t \in [0, 1]$ is the noise timestep. To compensate for the average weighting being halved, we increased $\lambda$ from 0.05 to 0.10. Table~\ref{tab:tcfm_ablation} shows results comparing CFM and TCFM.

\begin{table}[h]
\centering
\caption{CFM vs TCFM ablation at 400K iterations. TCFM does not improve upon standard CFM.}
\label{tab:tcfm_ablation}
\small
\begin{tabular}{l@{\hspace{1em}}c@{\hspace{1em}}c@{\hspace{1em}}c@{\hspace{1em}}c@{\hspace{1em}}c@{\hspace{1em}}c}
\toprule
Method & FID$\downarrow$ & sFID$\downarrow$ & IS$\uparrow$ & Prec.$\uparrow$ & Rec.$\uparrow$ & KDD$\downarrow$ \\
\midrule
CFM ($\lambda=0.05$) & \textbf{3.61} & \textbf{5.26} & \textbf{211.9} & \textbf{0.817} & \textbf{0.536} & \textbf{0.319} \\
TCFM ($\lambda=0.10$) & 3.82 & 5.57 & 211.8 & 0.813 & 0.533 & 0.321 \\
\bottomrule
\end{tabular}
\end{table}

Despite our hypothesis, TCFM underperformed standard CFM across most metrics. The time-weighting did not provide the expected benefit, suggesting that CFM's contrastive signal remains useful even at low noise levels.

\textbf{$x_0$ prediction with velocity loss.} We tested the approach from JiT~\cite{jit}, which uses $x_0$ prediction combined with velocity-based loss formulation. This technique showed promise for pixel-space diffusion models. However, in our latent space setting with INVAE, performance was significantly worse than standard flow matching. This suggests that $x_0$ prediction objectives may be most beneficial for pixel-space models rather than latent diffusion.

\textbf{Equilibrium Matching.} We attempted to implement Equilibrium Matching~\cite{eqm}, which combines energy-based modeling with flow matching. Despite following the methodology, we were unable to reproduce their reported results on our SiT-B/1 architecture with representation alignment. Performance was significantly worse than standard flow matching, suggesting potential incompatibilities between EqM and our architectural choices or training setup.

\subsection{Alternative Optimizers}

We evaluated alternative optimizers to determine if they could improve upon Adam's performance. Specifically, we tested Prodigy~\cite{prodigy}, which provides adaptive learning rate scheduling, and Muon~\cite{muon}, a momentum-based optimizer. Table~\ref{tab:optimizer_ablations} shows results at 400K iterations.

\begin{table}[h]
\centering
\caption{Alternative optimizer ablations at 400K iterations. All experiments use REG + INVAE + SPRINT + RMSNorm + RoPE. Adam with standard hyperparameters performs best.}
\label{tab:optimizer_ablations}
\small
\begin{tabular}{l@{\hspace{1em}}c@{\hspace{1em}}c@{\hspace{1em}}c@{\hspace{1em}}c@{\hspace{1em}}c@{\hspace{1em}}c}
\toprule
Optimizer & FID$\downarrow$ & sFID$\downarrow$ & IS$\uparrow$ & Prec.$\uparrow$ & Rec.$\uparrow$ & KDD$\downarrow$ \\
\midrule
Adam (baseline) & \textbf{4.09} &6.22 & \textbf{194.2} & 0.778 & \textbf{0.563} & 0.368 \\
Prodigy & 4.67 &  \textbf{5.70} & 175.1 &  \textbf{0.820} & 0.495 & \textbf{0.367} \\
Muon & 48.70 & 23.34 & 32.7 & 0.425 & 0.506 & 1.378 \\
\bottomrule
\end{tabular}
\end{table}

Both Prodigy and Muon showed early promise during training with lower initial losses. However, Adam consistently overtook both by 400K iterations. Muon's performance was particularly poor, plateauing early in training and achieving catastrophically bad final metrics (FID 48.70, KDD 1.378). Prodigy performed better than Muon but still underperformed Adam, particularly on recall (0.495 vs 0.563) and IS (175.1 vs 194.2). These results suggest that for our configuration, standard Adam with well-tuned hyperparameters remains the most reliable choice.

\section*{Visual Results}

All class-conditional samples below are generated with 250 sampling steps using path-drop guidance with guidance scale 2.5, 
guidance low threshold 0.10, and guidance high threshold 0.80.

\begin{figure}[p]
\centering
\setlength{\tabcolsep}{6pt}
\renewcommand{\arraystretch}{1.1}
\begin{tabular}{cc}
  \multicolumn{2}{c}{\textbf{Class 68}}\\
  \includegraphics[width=0.5\textwidth]{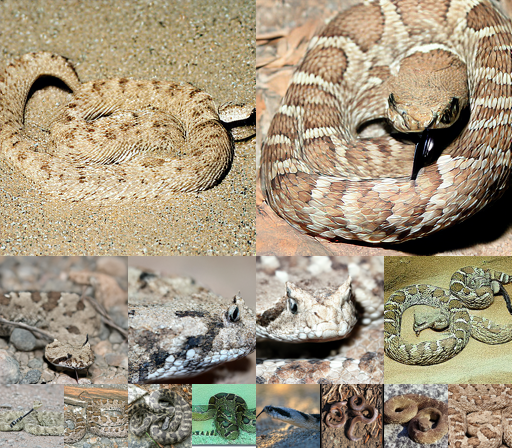} &
  \includegraphics[width=0.5\textwidth]{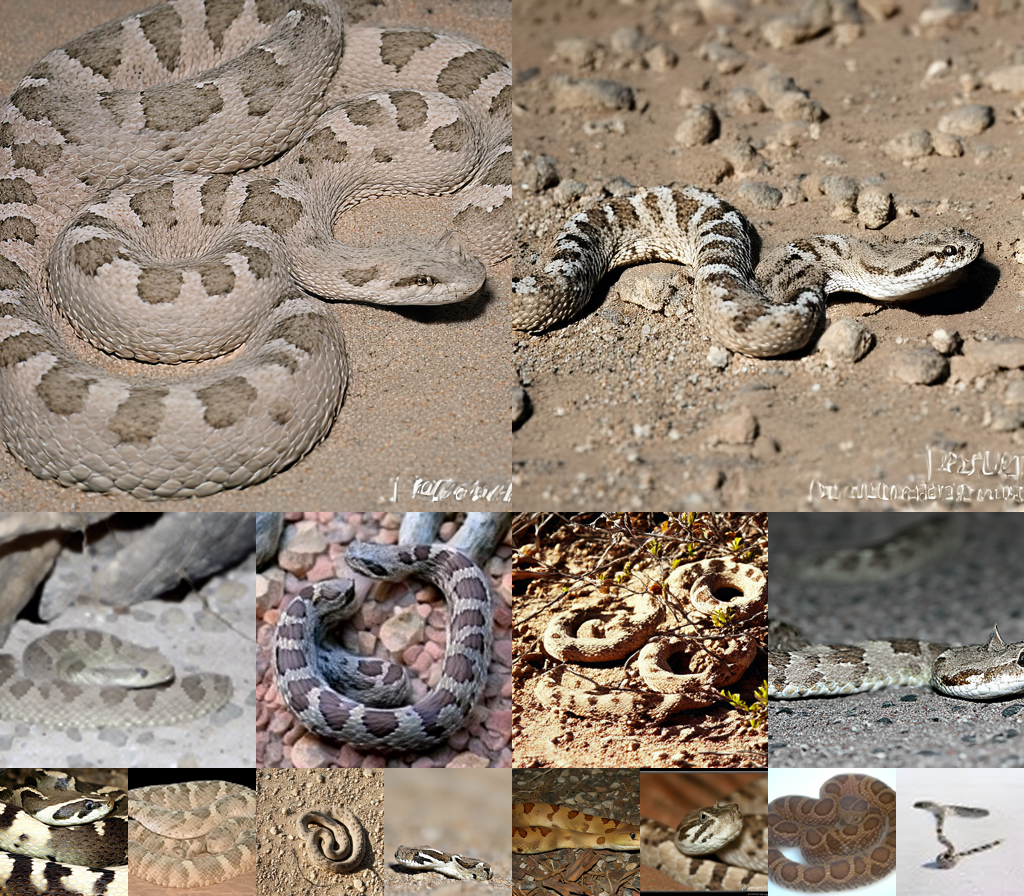} \\
  \scriptsize 256$\times$256 samples (ID 68) &
  \scriptsize 512$\times$512 samples (ID 68) \\
\end{tabular}
\caption{Uncurated SR-DiT-B/1 samples for ImageNet class 68 (sidewinder, horned rattlesnake, Crotalus cerastes) at 256$\times$256 and 512$\times$512 resolution.}
\label{fig:uncurated_label_samples_68}
\end{figure}

\begin{figure}[p]
\centering
\setlength{\tabcolsep}{6pt}
\renewcommand{\arraystretch}{1.1}
\begin{tabular}{cc}
  \multicolumn{2}{c}{\textbf{Class 92}}\\
  \includegraphics[width=0.5\textwidth]{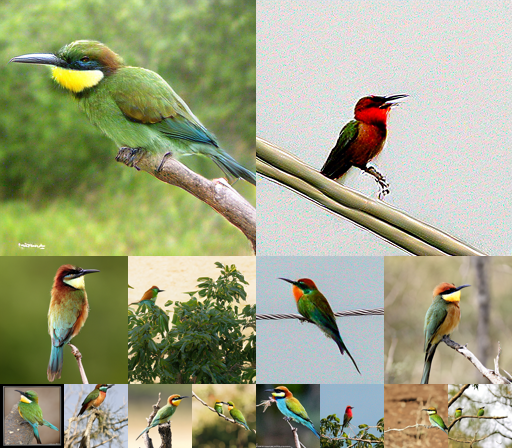} &
  \includegraphics[width=0.5\textwidth]{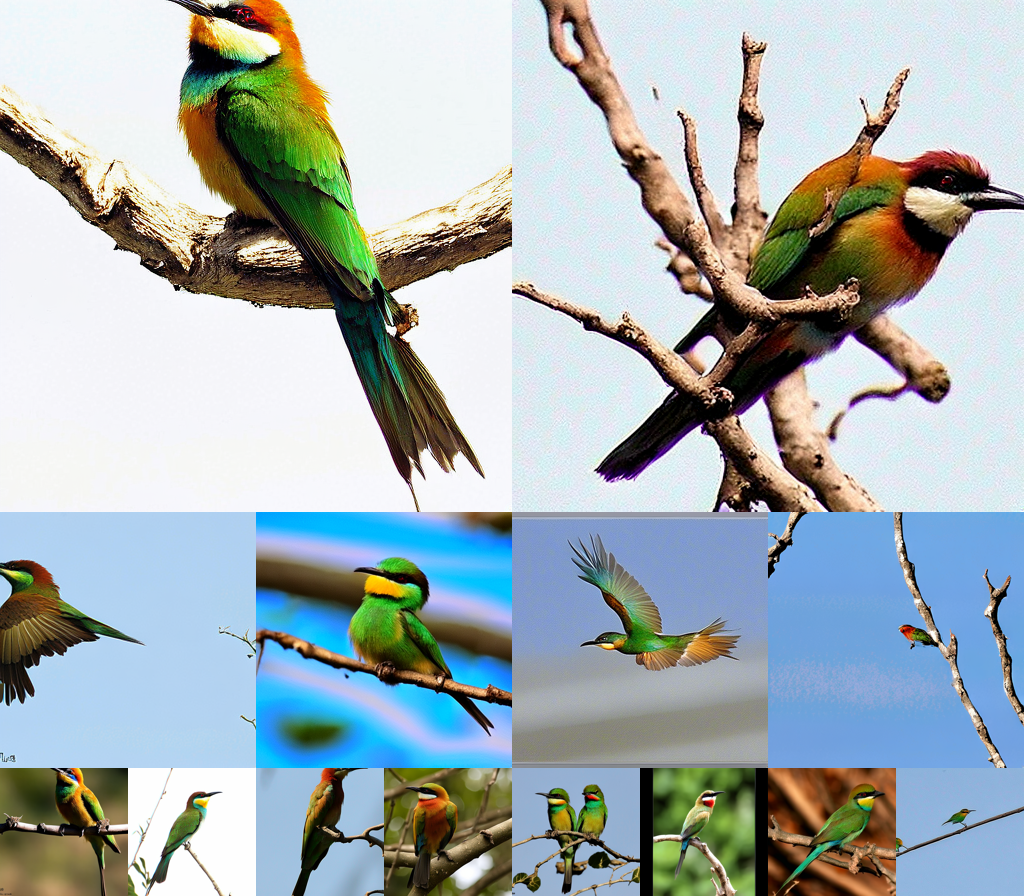} \\
  \scriptsize 256$\times$256 samples (ID 92) &
  \scriptsize 512$\times$512 samples (ID 92) \\
\end{tabular}
\caption{Uncurated SR-DiT-B/1 samples for ImageNet class 92 (bee eater) at 256$\times$256 and 512$\times$512 resolution.}
\label{fig:uncurated_label_samples_92}
\end{figure}

\begin{figure}[p]
\centering
\setlength{\tabcolsep}{6pt}
\renewcommand{\arraystretch}{1.1}
\begin{tabular}{cc}
  \multicolumn{2}{c}{\textbf{Class 233}}\\
  \includegraphics[width=0.5\textwidth]{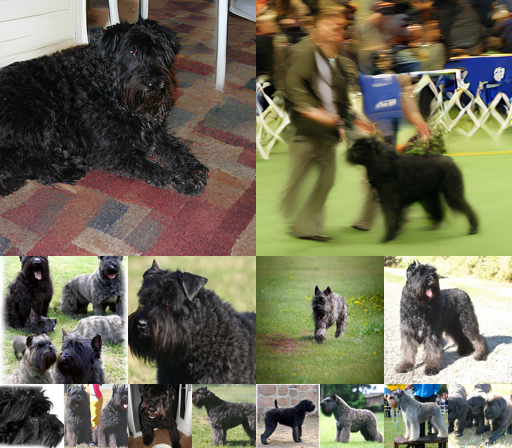} &
  \includegraphics[width=0.5\textwidth]{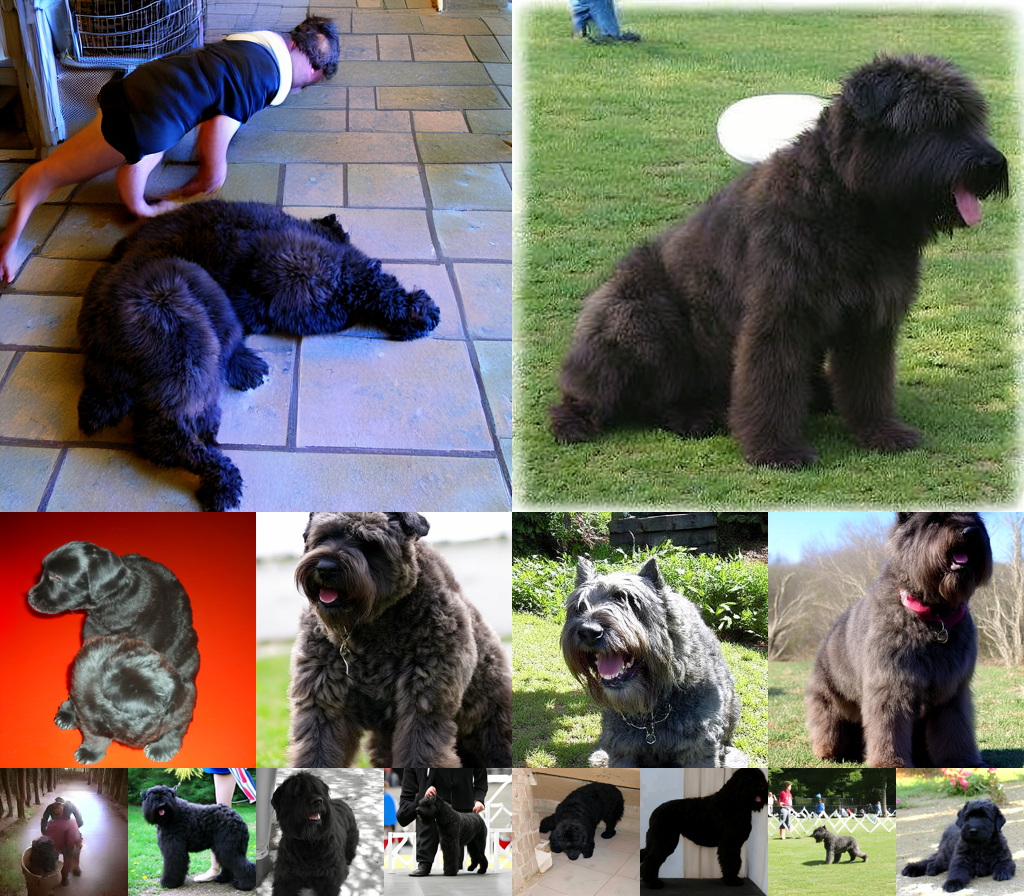} \\
  \scriptsize 256$\times$256 samples (ID 233) &
  \scriptsize 512$\times$512 samples (ID 233) \\
\end{tabular}
\caption{Uncurated SR-DiT-B/1 samples for ImageNet class 233 (Bouvier des Flandres, Bouviers des Flandres) at 256$\times$256 and 512$\times$512 resolution.}
\label{fig:uncurated_label_samples_233}
\end{figure}

\begin{figure}[p]
\centering
\setlength{\tabcolsep}{6pt}
\renewcommand{\arraystretch}{1.1}
\begin{tabular}{cc}
  \multicolumn{2}{c}{\textbf{Class 273}}\\
  \includegraphics[width=0.5\textwidth]{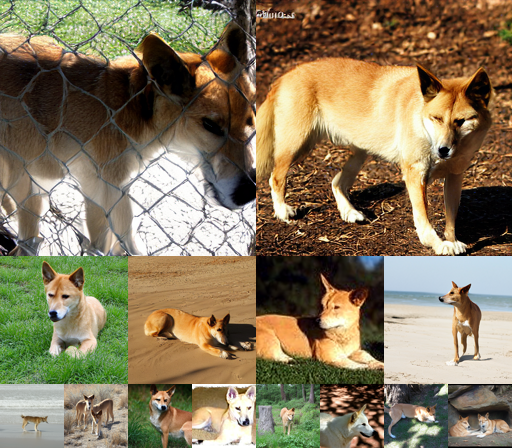} &
  \includegraphics[width=0.5\textwidth]{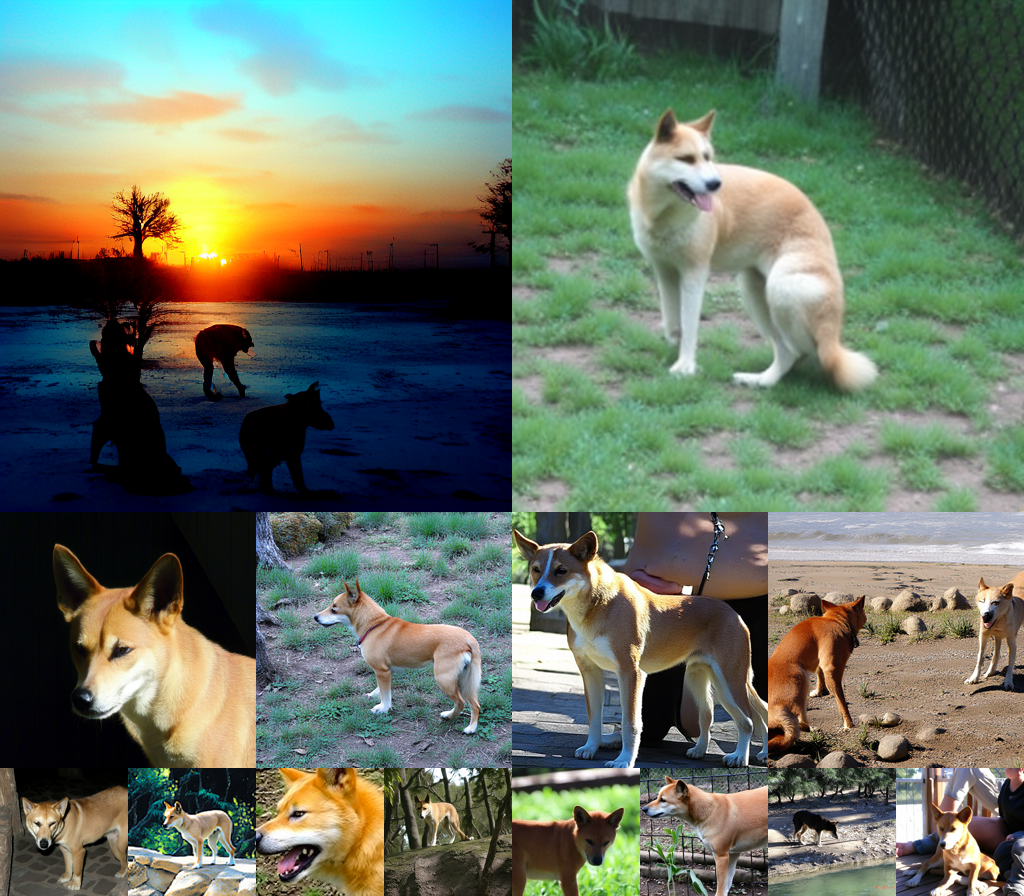} \\
  \scriptsize 256$\times$256 samples (ID 273) &
  \scriptsize 512$\times$512 samples (ID 273) \\
\end{tabular}
\caption{Uncurated SR-DiT-B/1 samples for ImageNet class 273 (dingo, warrigal, warragal, Canis dingo) at 256$\times$256 and 512$\times$512 resolution.}
\label{fig:uncurated_label_samples_273}
\end{figure}

\begin{figure}[p]
\centering
\setlength{\tabcolsep}{6pt}
\renewcommand{\arraystretch}{1.1}
\begin{tabular}{cc}
  \multicolumn{2}{c}{\textbf{Class 283}}\\
  \includegraphics[width=0.5\textwidth]{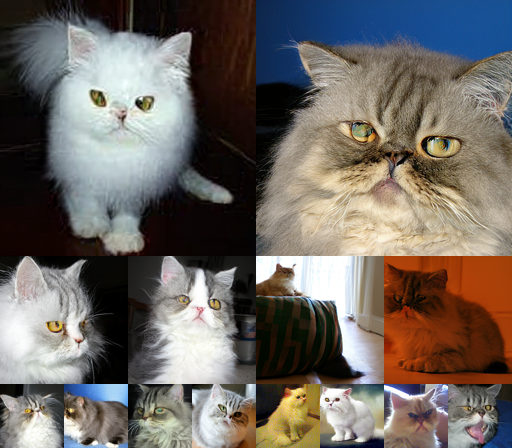} &
  \includegraphics[width=0.5\textwidth]{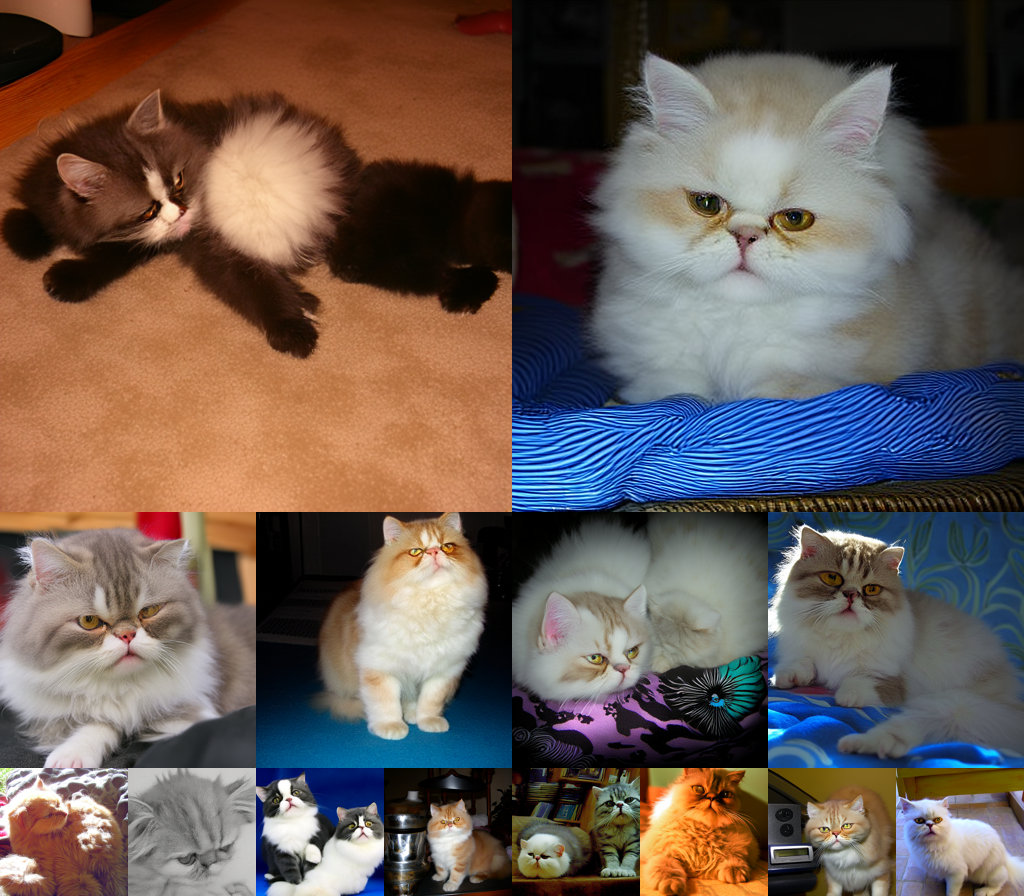} \\
  \scriptsize 256$\times$256 samples (ID 283) &
  \scriptsize 512$\times$512 samples (ID 283) \\
\end{tabular}
\caption{Uncurated SR-DiT-B/1 samples for ImageNet class 283 (Persian cat) at 256$\times$256 and 512$\times$512 resolution.}
\label{fig:uncurated_label_samples_283}
\end{figure}

\begin{figure}[p]
\centering
\setlength{\tabcolsep}{6pt}
\renewcommand{\arraystretch}{1.1}
\begin{tabular}{cc}
  \multicolumn{2}{c}{\textbf{Class 360}}\\
  \includegraphics[width=0.5\textwidth]{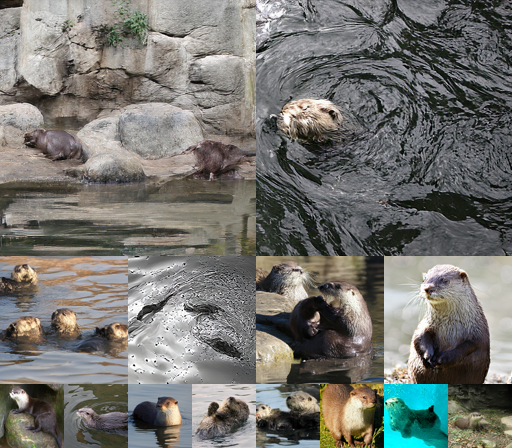} &
  \includegraphics[width=0.5\textwidth]{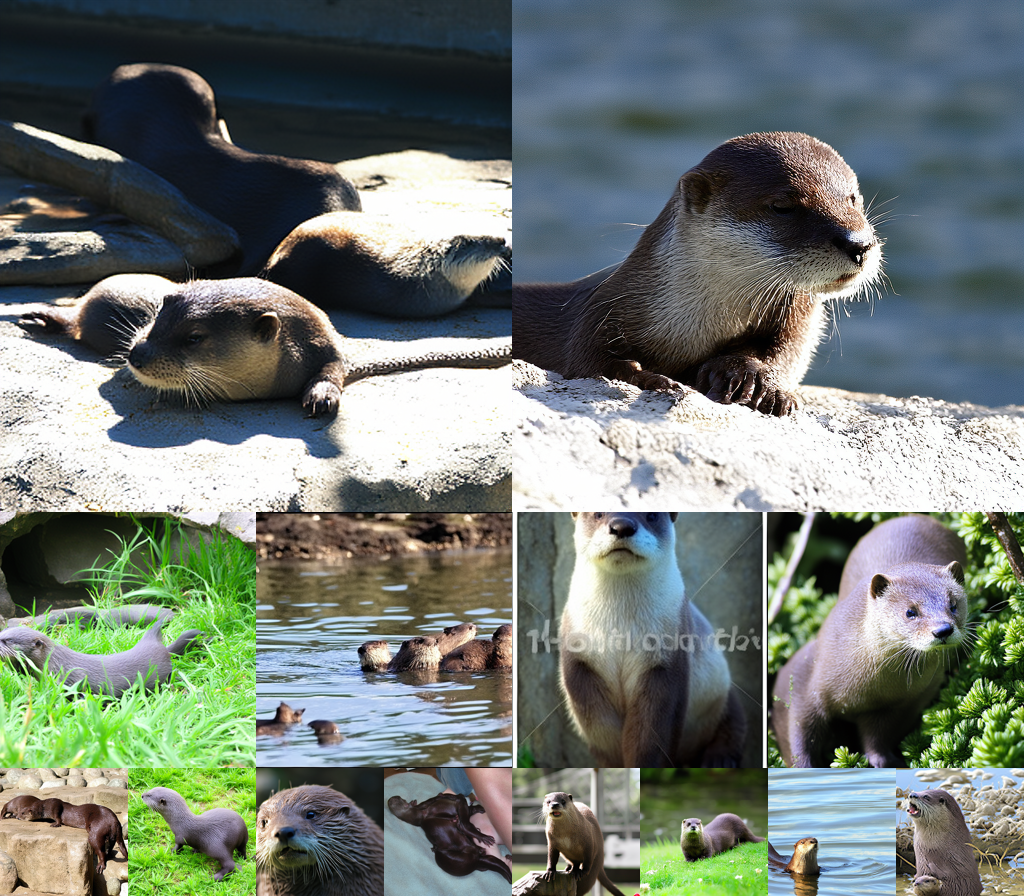} \\
  \scriptsize 256$\times$256 samples (ID 360) &
  \scriptsize 512$\times$512 samples (ID 360) \\
\end{tabular}
\caption{Uncurated SR-DiT-B/1 samples for ImageNet class 360 (otter) at 256$\times$256 and 512$\times$512 resolution.}
\label{fig:uncurated_label_samples_360}
\end{figure}

\begin{figure}[p]
\centering
\setlength{\tabcolsep}{6pt}
\renewcommand{\arraystretch}{1.1}
\begin{tabular}{cc}
  \multicolumn{2}{c}{\textbf{Class 482}}\\
  \includegraphics[width=0.5\textwidth]{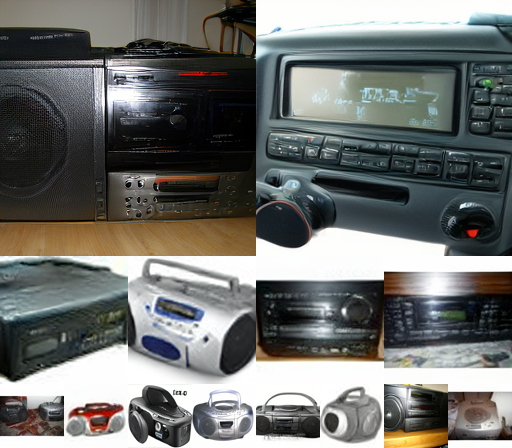} &
  \includegraphics[width=0.5\textwidth]{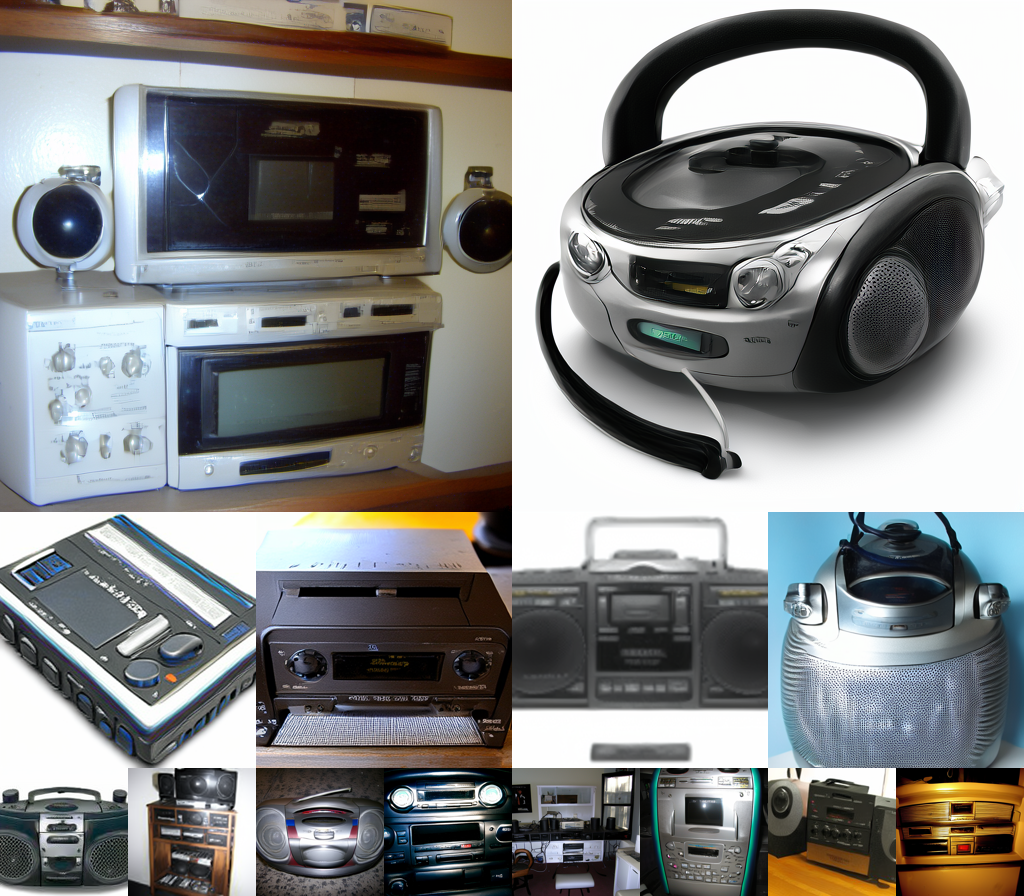} \\
  \scriptsize 256$\times$256 samples (ID 482) &
  \scriptsize 512$\times$512 samples (ID 482) \\
\end{tabular}
\caption{Uncurated SR-DiT-B/1 samples for ImageNet class 482 (cassette player) at 256$\times$256 and 512$\times$512 resolution.}
\label{fig:uncurated_label_samples_482}
\end{figure}

\begin{figure}[p]
\centering
\setlength{\tabcolsep}{6pt}
\renewcommand{\arraystretch}{1.1}
\begin{tabular}{cc}
  \multicolumn{2}{c}{\textbf{Class 545}}\\
  \includegraphics[width=0.5\textwidth]{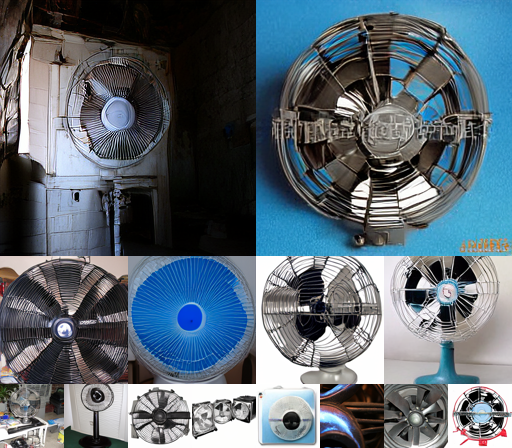} &
  \includegraphics[width=0.5\textwidth]{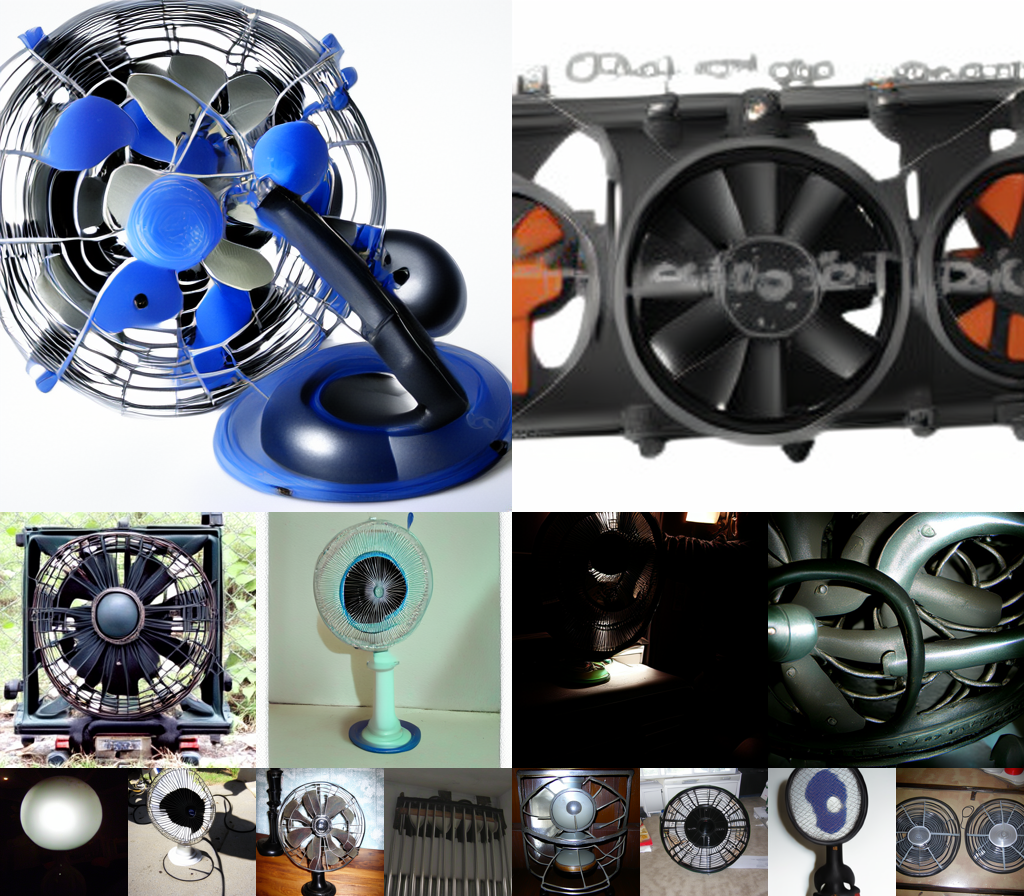} \\
  \scriptsize 256$\times$256 samples (ID 545) &
  \scriptsize 512$\times$512 samples (ID 545) \\
\end{tabular}
\caption{Uncurated SR-DiT-B/1 samples for ImageNet class 545 (electric fan, blower) at 256$\times$256 and 512$\times$512 resolution.}
\label{fig:uncurated_label_samples_545}
\end{figure}

\begin{figure}[p]
\centering
\setlength{\tabcolsep}{6pt}
\renewcommand{\arraystretch}{1.1}
\begin{tabular}{cc}
  \multicolumn{2}{c}{\textbf{Class 721}}\\
  \includegraphics[width=0.5\textwidth]{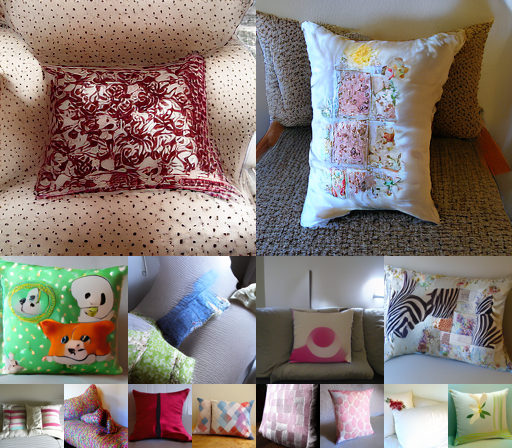} &
  \includegraphics[width=0.5\textwidth]{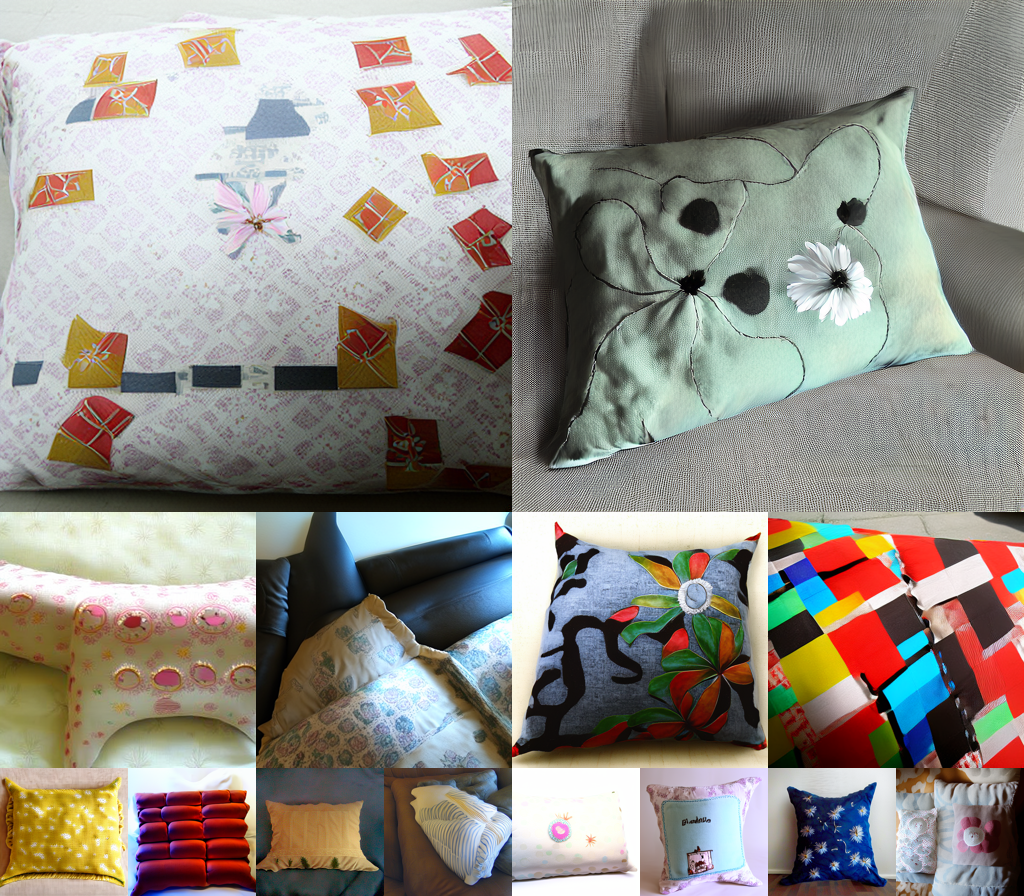} \\
  \scriptsize 256$\times$256 samples (ID 721) &
  \scriptsize 512$\times$512 samples (ID 721) \\
\end{tabular}
\caption{Uncurated SR-DiT samples for ImageNet class 721 (pillow) at 256$\times$256 and 512$\times$512 resolution.}
\label{fig:uncurated_label_samples_721}
\end{figure}

\begin{figure}[p]
\centering
\setlength{\tabcolsep}{6pt}
\renewcommand{\arraystretch}{1.1}
\begin{tabular}{cc}
  \multicolumn{2}{c}{\textbf{Class 727}}\\
  \includegraphics[width=0.5\textwidth]{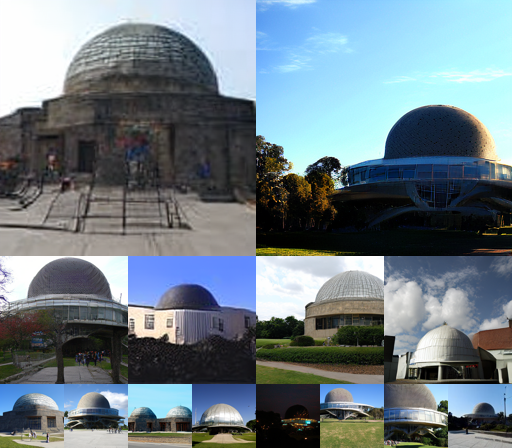} &
  \includegraphics[width=0.5\textwidth]{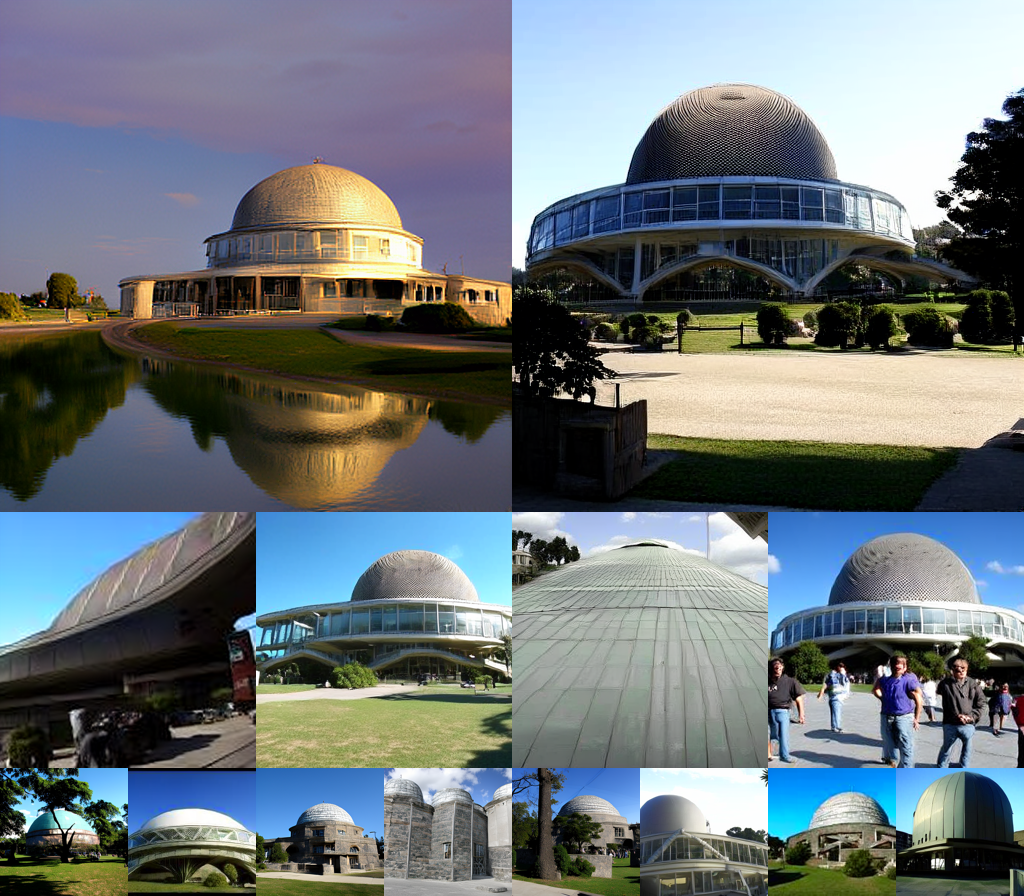} \\
  \scriptsize 256$\times$256 samples (ID 727) &
  \scriptsize 512$\times$512 samples (ID 727) \\
\end{tabular}
\caption{Uncurated SR-DiT samples for ImageNet class 727 (planetarium) at 256$\times$256 and 512$\times$512 resolution.}
\label{fig:uncurated_label_samples_727}
\end{figure}

\begin{figure}[p]
\centering
\setlength{\tabcolsep}{6pt}
\renewcommand{\arraystretch}{1.1}
\begin{tabular}{cc}
  \multicolumn{2}{c}{\textbf{Class 795}}\\
  \includegraphics[width=0.5\textwidth]{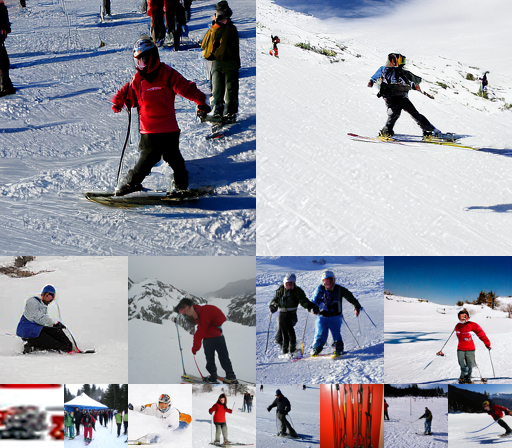} &
  \includegraphics[width=0.5\textwidth]{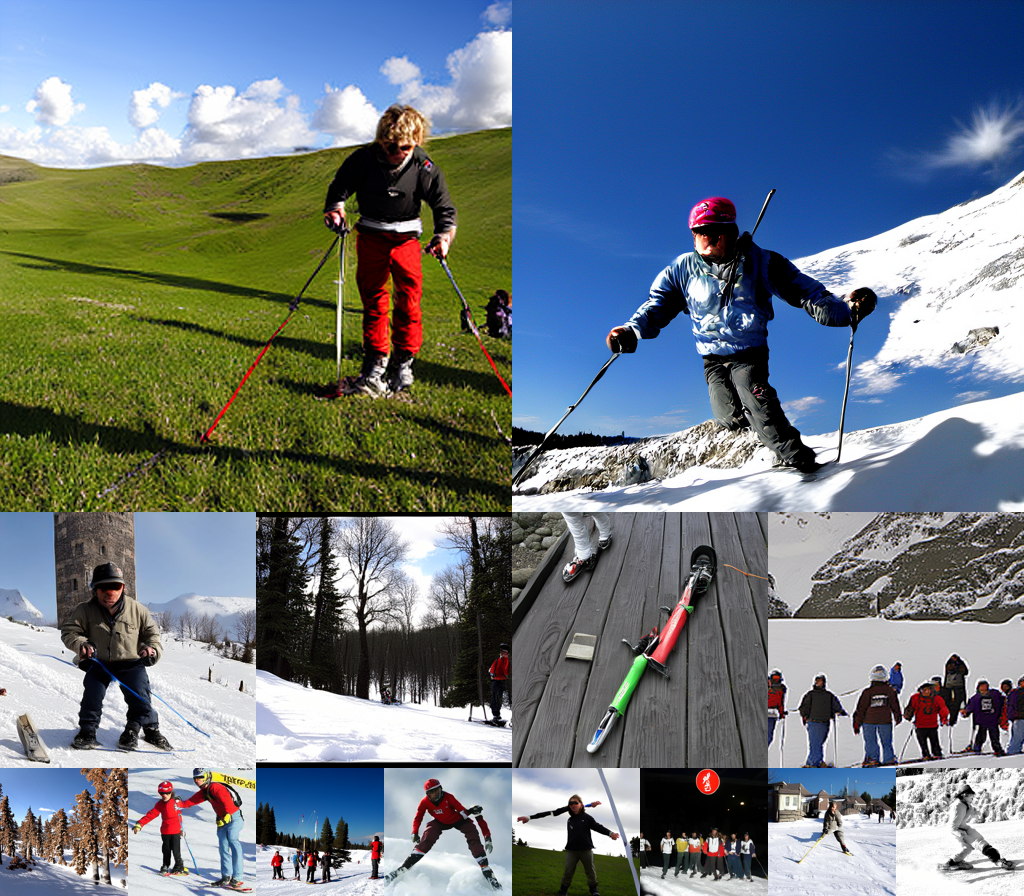} \\
  \scriptsize 256$\times$256 samples (ID 795) &
  \scriptsize 512$\times$512 samples (ID 795) \\
\end{tabular}
\caption{Uncurated SR-DiT samples for ImageNet class 795 (ski) at 256$\times$256 and 512$\times$512 resolution.}
\label{fig:uncurated_label_samples_795}
\end{figure}

\begin{figure}[p]
\centering
\setlength{\tabcolsep}{6pt}
\renewcommand{\arraystretch}{1.1}
\begin{tabular}{cc}
  \multicolumn{2}{c}{\textbf{Class 839}}\\
  \includegraphics[width=0.5\textwidth]{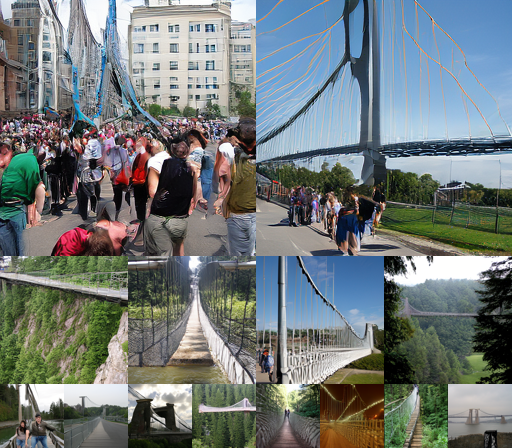} &
  \includegraphics[width=0.5\textwidth]{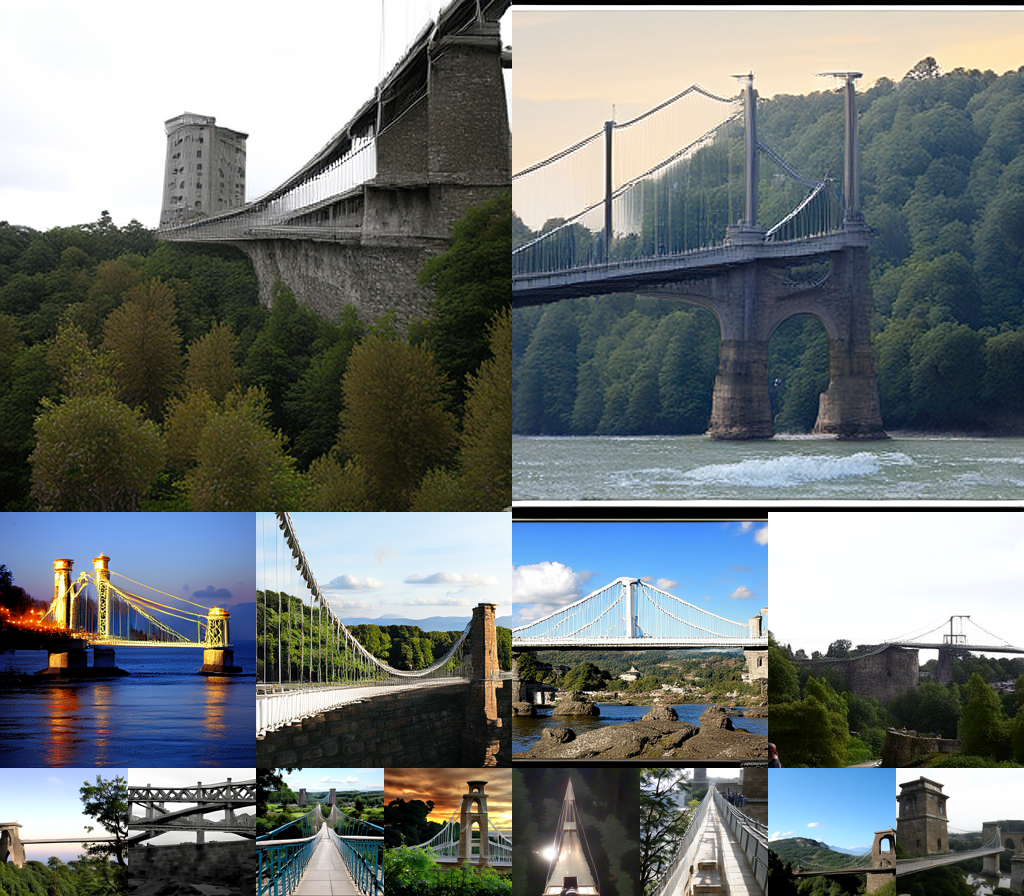} \\
  \scriptsize 256$\times$256 samples (ID 839) &
  \scriptsize 512$\times$512 samples (ID 839) \\
\end{tabular}
\caption{Uncurated SR-DiT samples for ImageNet class 839 (suspension bridge) at 256$\times$256 and 512$\times$512 resolution.}
\label{fig:uncurated_label_samples_839}
\end{figure}

\begin{figure}[p]
\centering
\setlength{\tabcolsep}{6pt}
\renewcommand{\arraystretch}{1.1}
\begin{tabular}{cc}
  \multicolumn{2}{c}{\textbf{Class 863}}\\
  \includegraphics[width=0.5\textwidth]{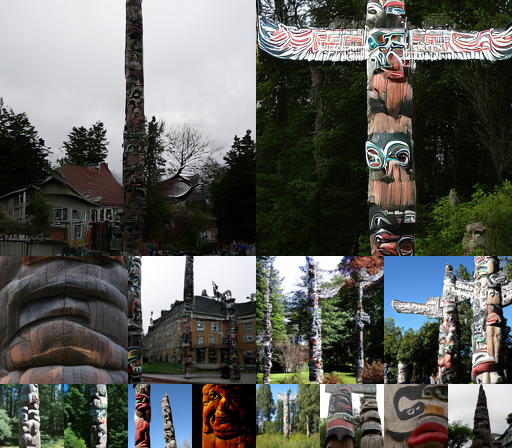} &
  \includegraphics[width=0.5\textwidth]{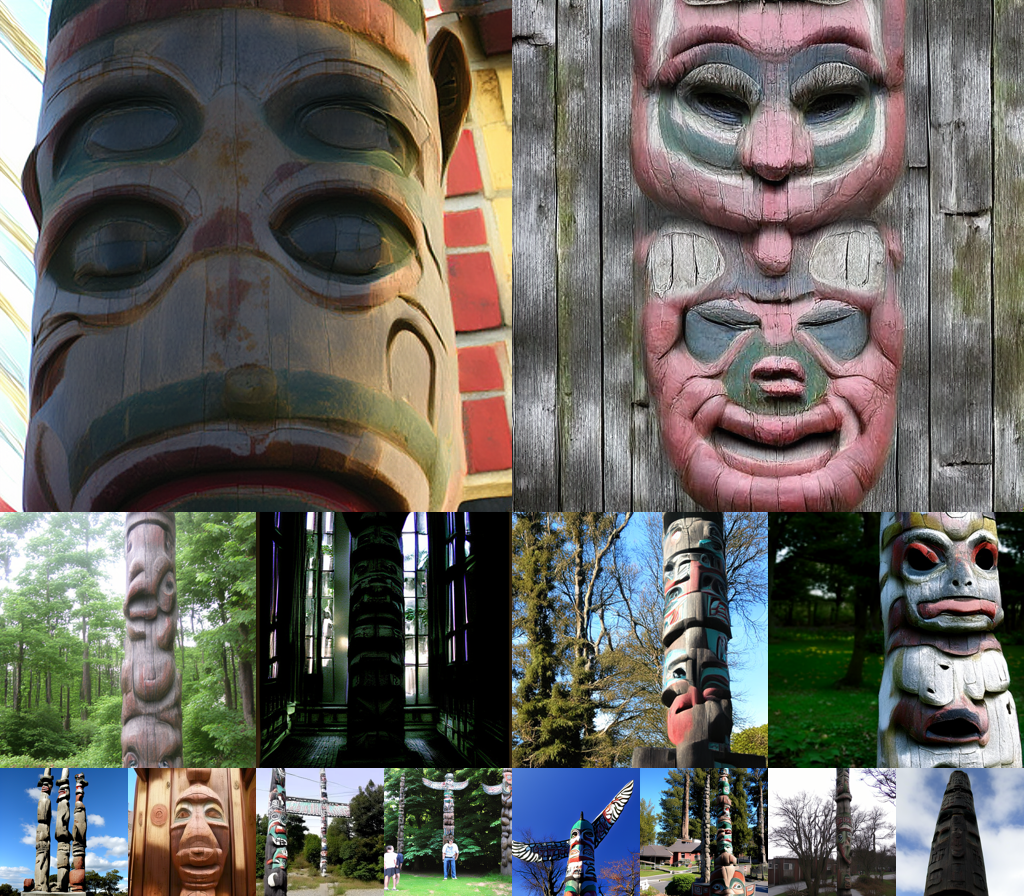} \\
  \scriptsize 256$\times$256 samples (ID 863) &
  \scriptsize 512$\times$512 samples (ID 863) \\
\end{tabular}
\caption{Uncurated SR-DiT samples for ImageNet class 863 (totem pole) at 256$\times$256 and 512$\times$512 resolution.}
\label{fig:uncurated_label_samples_863}
\end{figure}

\begin{figure}[p]
\centering
\setlength{\tabcolsep}{6pt}
\renewcommand{\arraystretch}{1.1}
\begin{tabular}{cc}
  \multicolumn{2}{c}{\textbf{Class 929}}\\
  \includegraphics[width=0.5\textwidth]{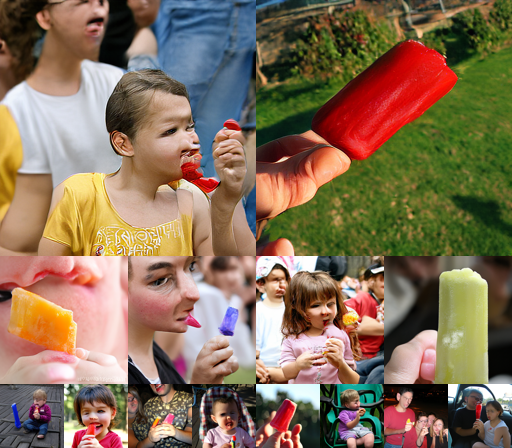} &
  \includegraphics[width=0.5\textwidth]{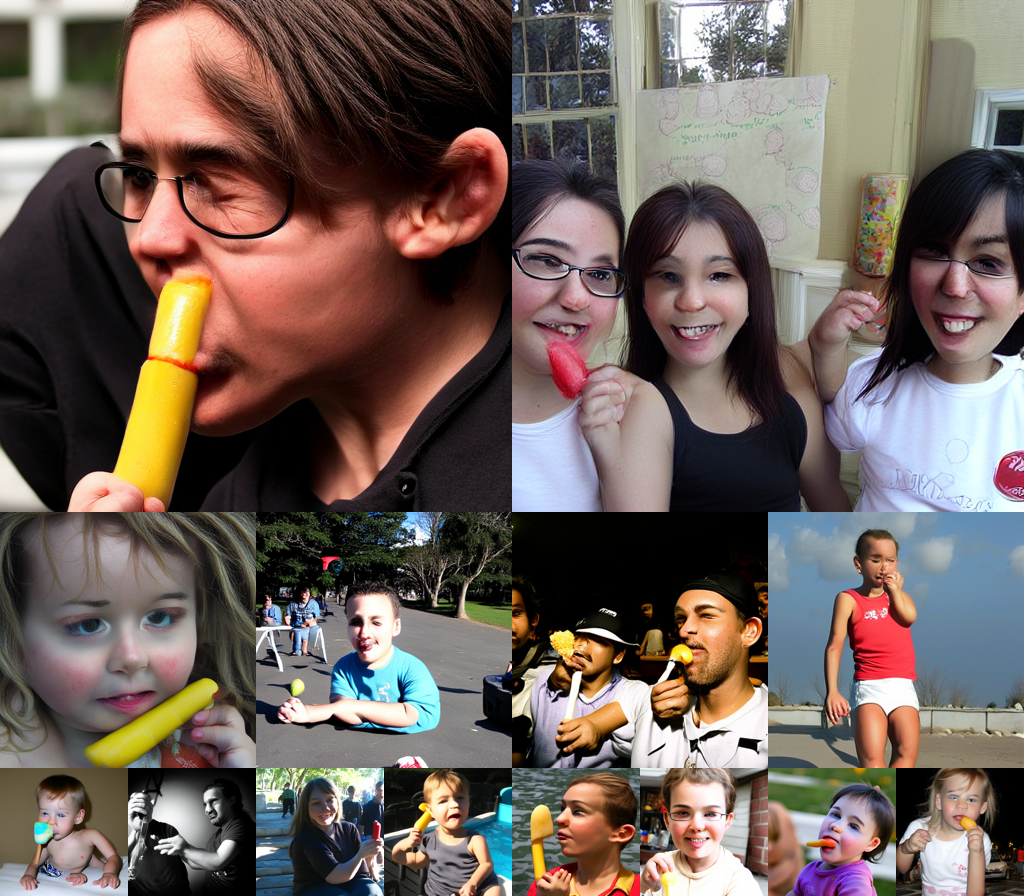} \\
  \scriptsize 256$\times$256 samples (ID 929) &
  \scriptsize 512$\times$512 samples (ID 929) \\
\end{tabular}
\caption{Uncurated SR-DiT samples for ImageNet class 929 (ice lolly, lolly, lollipop, popsicle) at 256$\times$256 and 512$\times$512 resolution.}
\label{fig:uncurated_label_samples_929}
\end{figure}

\begin{figure}[p]
\centering
\setlength{\tabcolsep}{6pt}
\renewcommand{\arraystretch}{1.1}
\begin{tabular}{cc}
  \multicolumn{2}{c}{\textbf{Class 944}}\\
  \includegraphics[width=0.5\textwidth]{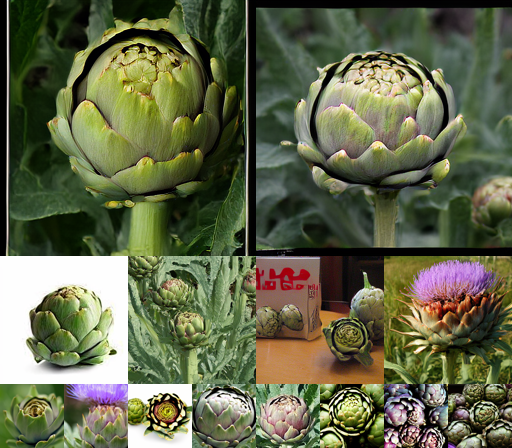} &
  \includegraphics[width=0.5\textwidth]{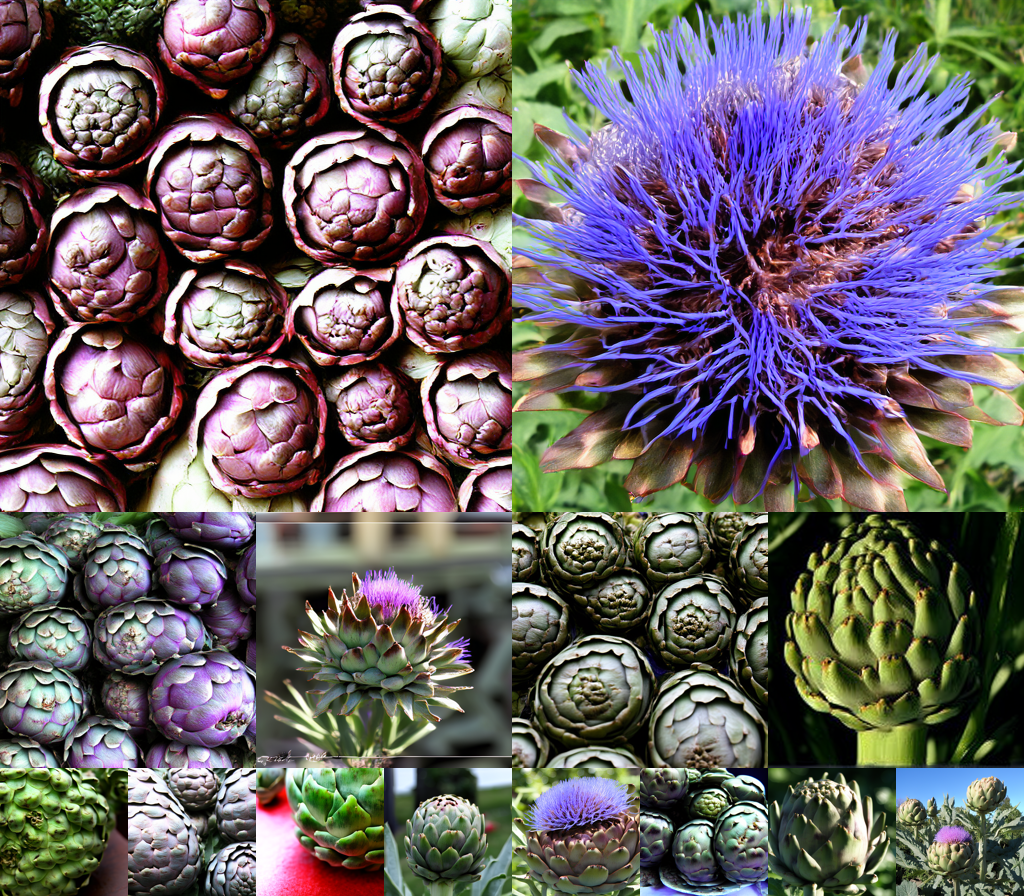} \\
  \scriptsize 256$\times$256 samples (ID 944) &
  \scriptsize 512$\times$512 samples (ID 944) \\
\end{tabular}
\caption{Uncurated SR-DiT samples for ImageNet class 944 (artichoke, globe artichoke) at 256$\times$256 and 512$\times$512 resolution.}
\label{fig:uncurated_label_samples_944}
\end{figure}

\begin{figure}[p]
\centering
\setlength{\tabcolsep}{6pt}
\renewcommand{\arraystretch}{1.1}
\begin{tabular}{cc}
  \multicolumn{2}{c}{\textbf{Class 974}}\\
  \includegraphics[width=0.5\textwidth]{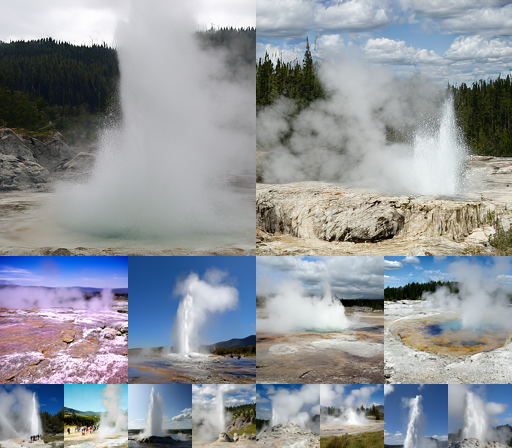} &
  \includegraphics[width=0.5\textwidth]{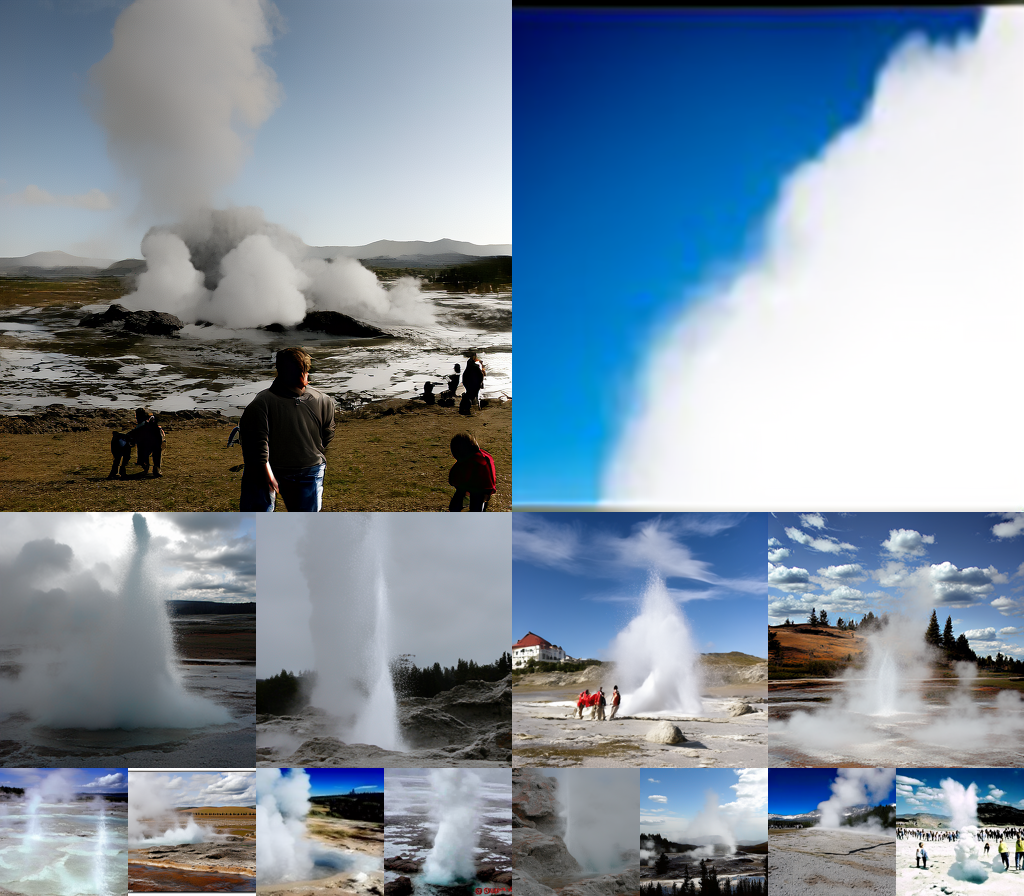} \\
  \scriptsize 256$\times$256 samples (ID 974) &
  \scriptsize 512$\times$512 samples (ID 974) \\
\end{tabular}
\caption{Uncurated SR-DiT samples for ImageNet class 974 (geyser) at 256$\times$256 and 512$\times$512 resolution.}
\label{fig:uncurated_label_samples_974}
\end{figure}

\begin{figure}[p]
\centering
\setlength{\tabcolsep}{6pt}
\renewcommand{\arraystretch}{1.1}
\begin{tabular}{cc}
  \multicolumn{2}{c}{\textbf{Class 984}}\\
  \includegraphics[width=0.5\textwidth]{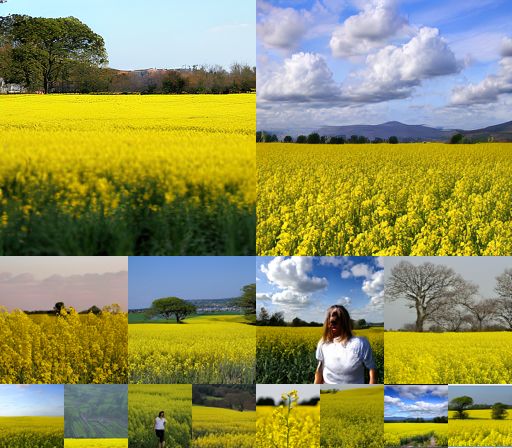} &
  \includegraphics[width=0.5\textwidth]{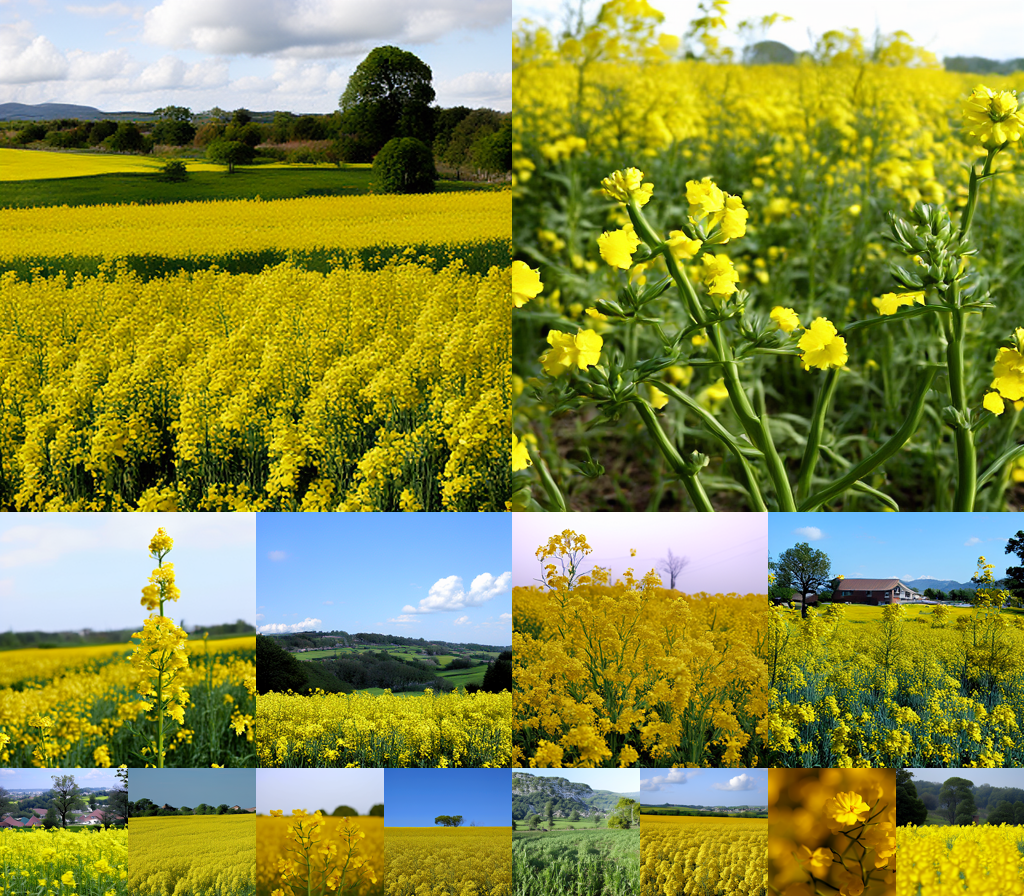} \\
  \scriptsize 256$\times$256 samples (ID 984) &
  \scriptsize 512$\times$512 samples (ID 984) \\
\end{tabular}
\caption{Uncurated SR-DiT samples for ImageNet class 984 (rapeseed) at 256$\times$256 and 512$\times$512 resolution.}
\label{fig:uncurated_label_samples_984}
\end{figure}

\begin{figure}[p]
\centering
\setlength{\tabcolsep}{6pt}
\renewcommand{\arraystretch}{1.1}
\begin{tabular}{cc}
  \multicolumn{2}{c}{\textbf{Class 991}}\\
  \includegraphics[width=0.5\textwidth]{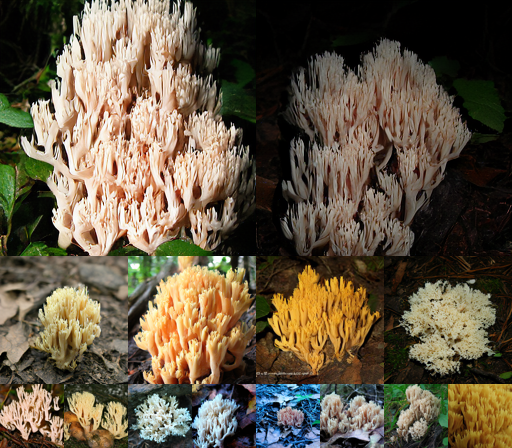} &
  \includegraphics[width=0.5\textwidth]{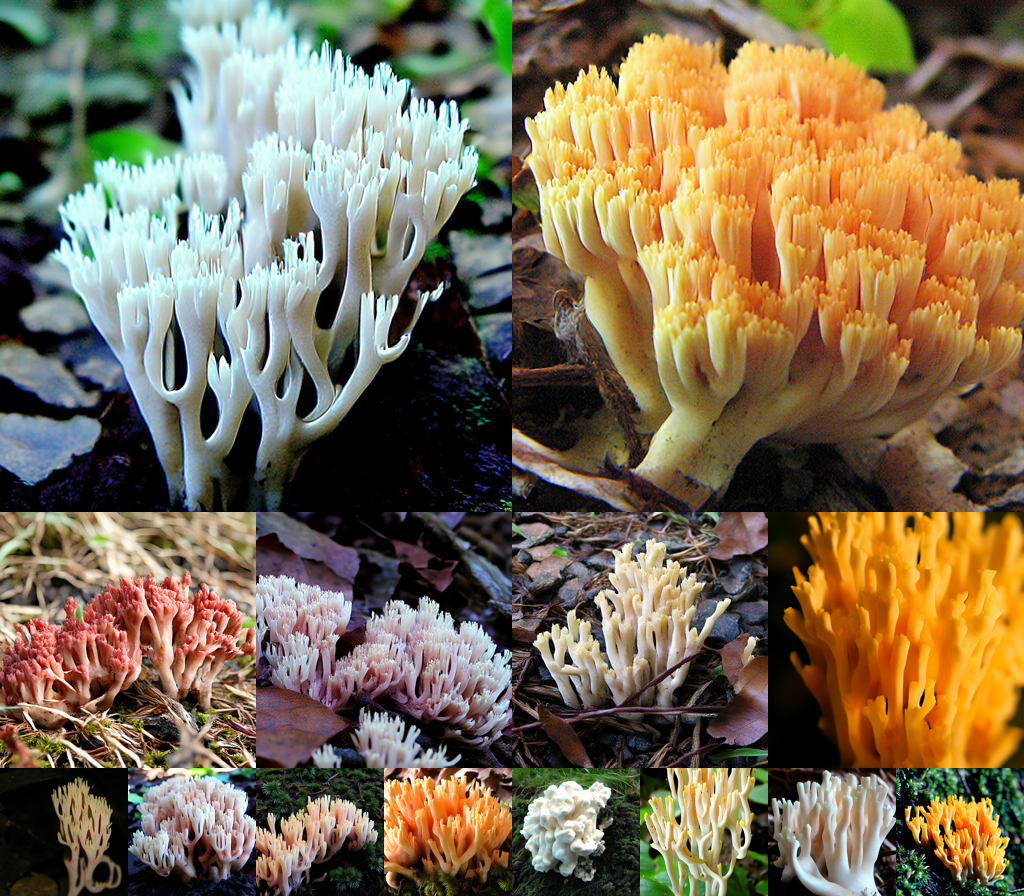} \\
  \scriptsize 256$\times$256 samples (ID 991) &
  \scriptsize 512$\times$512 samples (ID 991) \\
\end{tabular}
\caption{Uncurated SR-DiT samples for ImageNet class 991 (coral fungus) at 256$\times$256 and 512$\times$512 resolution.}
\label{fig:uncurated_label_samples_991}
\end{figure}

\end{document}